%% file: main.tex
\documentclass[10pt,twocolumn,letterpaper]{article}

\usepackage{cvpr}              
\usepackage{xcolor}    
\usepackage{colortbl}
\usepackage{tcolorbox}
\usepackage{graphicx}


\renewcommand{\paragraph}[1]{\vspace{1.25mm}\noindent\textbf{#1}}
\newcommand{\blue}[1]{\textbf{\textcolor{mblue}{#1}}}
\definecolor{mblue}{RGB}{0, 77, 128}
\definecolor{cvprblue}{rgb}{0.21,0.49,0.74}
\usepackage[pagebackref,breaklinks,colorlinks,allcolors=cvprblue]{hyperref}
\usepackage{arydshln}
\usepackage{multirow}

\usepackage{pifont}
\def\paperID{10283} 
\def\confName{CVPR}
\def\confYear{2025}

\title{AIGV-Assessor: Benchmarking and Evaluating the Perceptual Quality of \\ Text-to-Video Generation with LMM}

\author{
Jiarui Wang$^{1}$, \qquad 
Huiyu Duan$^{1,2}$, \qquad 
Guangtao Zhai$^{1,2}$, \qquad 
Juntong Wang$^{1}$, \qquad 
Xiongkuo Min$^{1}$\thanks{Corresponding Author.},\qquad 
\\
$^{1}$Institute of Image Communication and Network Engineering, \\
$^{2}$ MoE Key Lab of Artificial Intelligence, AI Institute,\\ Shanghai Jiao Tong University, Shanghai, China\\}

\begin{document}
\maketitle
\input{section/0_abstract}
\input{section/1_introduction}
\input{section/2_related}

\input{section/3_database}

\input{section/4_model}

\input{section/5_exp}
\input{section/6_conclusion}

{
    \small
    \bibliographystyle{ieeenat_fullname}
    \bibliography{main.bib}
}
\newpage
\clearpage
\newpage
\appendix
\input{supple/0_importance}
\input{supple/1_datadetails}

\input{supple/2_subjective}

\input{supple/3_Method}

\input{supple/4_Experiment}

\input{supple/5_figures}

\end{document}

%% file: section/0_abstract.tex
\begin{abstract}
The rapid advancement of large multimodal models (LMMs) has led to the rapid expansion of artificial intelligence generated videos (AIGVs), which highlights the pressing need for effective video quality assessment (VQA) models designed specifically for AIGVs. 
Current VQA models generally fall short in accurately assessing the perceptual quality of AIGVs due to the presence of unique distortions, such as unrealistic objects, unnatural movements, or inconsistent visual elements. 
To address this challenge, we first present \textbf{AIGVQA-DB}, a large-scale dataset comprising 36,576 AIGVs generated by 15 advanced text-to-video models using 1,048 diverse prompts. 
With these AIGVs, a systematic annotation pipeline including scoring and ranking processes is devised, which collects 370k expert ratings to date.
Based on AIGVQA-DB, we further introduce \textbf{AIGV-Assessor}, a novel VQA model that leverages spatiotemporal features and LMM frameworks to capture the intricate quality attributes of AIGVs, thereby accurately predicting precise video quality scores and video pair preferences.
Through comprehensive experiments on both AIGVQA-DB and existing AIGV databases, AIGV-Assessor demonstrates state-of-the-art performance, significantly surpassing existing scoring or evaluation methods in terms of multiple perceptual quality dimensions. 
The dataset and code will be released at \url{https://github.com/wangjiarui153/AIGV-Assessor}.
\end{abstract}

%% file: section/1_introduction.tex
\section{Introduction}
\label{sec:intro}
\input{figure/annotation}
Text-to-video generative models~\cite{wu2023tune, khachatryan2023text2video, chen2023videocrafter1,luo2023videofusion,wang2023modelscope}, including auto-regressive \cite{yan2021videogpt,hong2022cogvideo} and diffusion-based \cite{singer2022make,khachatryan2023text2video,chen2023videocrafter1} approaches, have experienced rapid advancements in recent years with the explosion of large multimodal models (LMMs).
Given appropriate text prompts, these models can generate high-fidelity and semantically-aligned videos, commonly referred to as AI-generated videos (AIGVs), which have significantly facilitated the content creation in various domains, including entertainment, art, design, and advertising, \textit{etc} \cite{xu2024imagereward,liang2024rich}.
Despite the significant progress, current AIGVs are still far from satisfactory.
Unlike natural videos, which are usually affected by low-level distortions, such as noise, blur, low-light, \textit{etc}, AIGVs generally suffer from degradations such as unrealistic objects, unnatural movements, inconsistent visual elements, and misalignment with text descriptions \cite{liu2024fetv,huang2024vbench,zhang2024benchmarking,kou2024subjective}.

The unique distortions in AIGVs also bring challenges to the video evaluation.
Traditional video quality assessment (VQA) methods~\cite{VSFA, li2022blindly, sun2022a, wu2022fast, wu2023dover} mainly focus on evaluating the quality of professionally-generated content (PGC) and user-generated content (UGC), thus struggling to address the specific distortions associated with AIGVs, such as spatial artifacts, temporal inconsistencies, and misalignment between generated content and text prompts.
For evaluation of AIGVs, some metrics such as Inception Score (IS)~\cite{IS} and Fréchet Video Distance (FVD)~\cite{unterthiner2018towards} have been widely used, which are computed over distributions of videos and may not reflect the human preference for an individual video.
Moreover, these metrics mainly evaluate the fidelity of videos, while failing to assess the text-video correspondence.
Vision-language pre-training models, such as CLIPScore~\cite{CLIPScore}, BLIPScore~\cite{li2022blip}, and AestheticScore~\cite{schuhmann2022laion} are frequently employed to evaluate the alignment between generated videos and their text prompts.
However, these models mainly consider the text-video alignment at the image level, while ignoring the dynamic diversity and motion consistency of visual elements that are crucial to the video-viewing experience.

In this paper, to facilitate the development of more comprehensive and precise metrics for evaluating AI-generated videos, we present \textbf{AIGVQA-DB}, a large-scale VQA dataset, including 36,576 AIGVs generated by 15 advanced text-to-video models using 1,048 diverse prompts. 
An overview of the dataset construction pipeline is shown in Figure \ref{annotation}.
The prompts are collected from existing open-domain text-video datasets~\cite{liu2024fetv,videoworldsimulators2024, wang2023internvid, xu2016msr-vtt,Bain21,li2016tgif} or manually-written, which can be categorized based on four orthogonal aspects respectively, as shown in Figure \ref{annotation}(a)-(d). 
Based on the AIGVs, we collect 370k expert ratings comprising both mean opinion scores (MOSs) and pairwise comparisons, which are evaluated from four dimensions, including: (1) static quality, (2) temporal smoothness, (3) dynamic degree, and (4) text-video correspondence.
Equipped with the dataset, we propose \textbf{AIGV-Assessor}, a large multimodal model-based (LMM-based) VQA method for AIGVs, which reformulates the quality regression task into an interactive question-and-answer (Q\&A) framework and leverages the powerful multimodal representation capabilities of LMMs to provide accurate and robust quality assessments.
AIGV-Assessor not only classifies videos into different quality levels through natural language output, but also generates precise quality scores through regression, thus enhancing the interpretability and usability of VQA results.
Moreover, AIGV-Assessor also excels in pairwise video comparisons, enabling nuanced assessments that are closer to human preferences. 
Extensive experimental results demonstrate that AIGV-Assessor outperforms existing text-to-video scoring methods in terms of multiple dimensions relevant to human preference.
The main contributions of this paper are summarized as follows:
\begin{itemize}
\item We construct AIGVQA-DB, a large-scale dataset comprising 36,576 AI-generated videos annotated with MOS scores and pairwise comparisons. Compared with existing benchmarks, AIGVQA-DB provides a more comprehensive assessment of the capabilities of text-to-video models from multiple perspectives.

\item Based on AIGVQA-DB, we evaluate and benchmark 15 representative text-to-video models, and reveal their strengths and weaknesses from four crucial preference dimensions, \textit{i.e.,} static quality, temporal smoothness, dynamic degree, and text-to-video correspondence.


\item We present a novel LMM-based VQA model for AIGVs, termed AIGV-Assessor, which integrates both spatial and temporal visual features as well as prompt features into a LMM to give quality levels, predict quality scores, and conduct quality comparisions.


\item Thorough analysis of our AIGV-Assessor is provided and extensive experiments on our proposed AIGVQA-DB and other AIGV quality assessment datasets have shown the effectiveness and applicability of AIGV-Assessor.


\end{itemize}

%% file: figure/annotation.tex
\begin{figure*}[!t]
\vspace{-6mm}
	\centering
	\includegraphics[width=\linewidth]{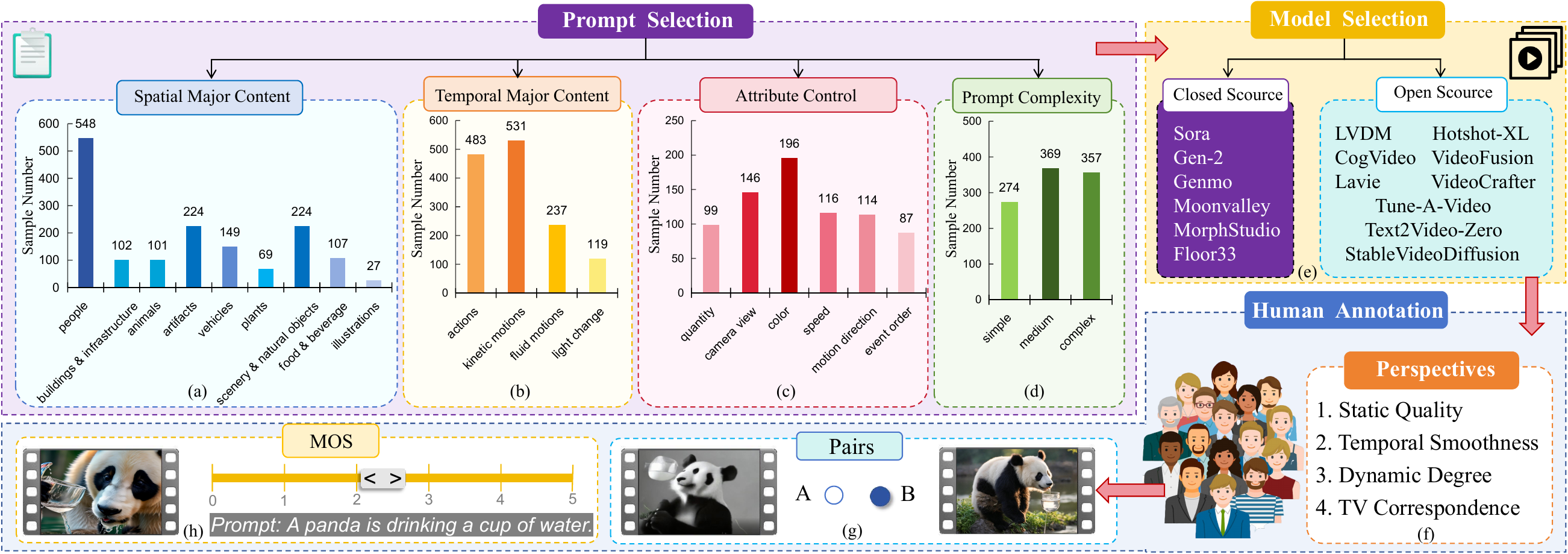}
 \vspace{-7mm}
	\caption{An overview of the AIGVQA-DB construction pipeline, illustrating the generation and the subjective evaluation procedures for the AIGVs in the database. (a) Prompt categorization according to the spatial major content. (b) Prompt categorization according to the temporal descriptions. (c) Prompt categorization according to the attribute control. (d) Prompt categorization according to the prompt complexity. (e) The 15 generative models used in the database. (f) Four visual quality evaluation perspectives, including static quality, temporal smoothness, dynamic degree, and text-video correspondence. (g) and (h) demonstrates the pair comparison and preference scoring processes, respectively. }
 \vspace{-4mm}
	\label{annotation}
\end{figure*}

%% file: section/2_related.tex
\vspace{-1mm}
\section{Related Work}
\label{sec:related}
\vspace{-1mm}
\input{table/models}
\subsection{Text-to-video Generation}
\input{table/database}
\vspace{-1mm}
Recent advancements in text-to-video generative models have substantially broadened video creation and modification possibilities. 
As shown in Table \ref{Video_model}, these models exhibit distinct characteristics and capacities, including modes, resolution, and total frames. 
CogVideo \cite{hong2022cogvideo} is an early text-to-video (T2V) model capable of generating short videos based on CogView2~\cite{ding2022cogview2}. 
Make-a-video~\cite{singer2022make} adds effective spatial-temporal modules on a diffusion-based text-to-image (T2I) model (\ie, DALLE-2~\cite{ramesh2022hierarchical}). VideoFusion~\cite{luo2023videofusion} also leverages the DALLE-2 and presents a decomposed diffusion process. 
LVDM~\cite{he2022latent}, Text2Video-Zero~\cite{khachatryan2023text2video}, Tune-A-Video~\cite{wu2023tune}, and ModelScope~\cite{wang2023modelscope} are models that inherit the success of Stable Diffusion (SD) ~\cite{rombach2022high} for video generation. 
Lavie~\cite{wang2023lavie} extends the original transformer block in SD to a spatio-temporal transformer. Hotshot-XL \cite{Mullan_Hotshot-XL_2023} introduces personalized video generation.
Beyond these laboratory-driven advancements, the video generation landscape has also been enriched by a series of commercial products. Notable among them are Floor33 \cite{floor33}, Gen-2 \cite{gen}, Genmo \cite{gem},  MoonValley \cite{moonvalley}, MorphStudio  \cite{morph},  and Sora~\cite{videoworldsimulators2024}, which have gained substantial attention in both academia and industry, demonstrating the widespread application potential of AI-assisted video creation. 

\vspace{-1mm}
\subsection{Text-to-video Evaluation}
\vspace{-1mm}
The establishment of the AI-generated image quality assessment (AIGIQA) dataset is relatively well-developed, including both mean opinion scores (MOSs) for absolute quality evaluations, and pairwise comparisons for relative quality judgments. 
Recent developments in text-to-video generation models have also spurred the creation of various AI-generated video quality assessment (AIGVQA) datasets, addressing different aspects of the T2V generation challenge, as shown in Table \ref{tab:dataset}. 
MQT~\cite{chivileva2023measuring} consists of 1,005 videos generated by 5 models using 201 prompts.
EvalCrafter~\cite{liu2023evalcrafter} and FETV~\cite{liu2024fetv} extend the scale of the videos, prompts, and evaluation dimensions. 
LGVQ~\cite{zhang2024benchmarking} increases the number of annotators, providing more reliable MOSs.
T2VQA-DB~\cite{kou2024subjective} consists of 10,000 videos from 1,000 prompts representing a significant improvement in scale.
GAIA~\cite{chen2024gaia} collects 9,180 videos focusing on action quality assessment in AIGVs, but falls short in addressing the consistency between the generated visuals and their textual prompts. 
Most existing VQA datasets predominantly rely on MOS, an absolute scoring method, which suffers from the same drawback: absolute scores alone may cause ambiguity and overlook subtle quality differences.
In contrast, our AIGVQA-DB includes both MOSs and pairwise comparisons, addressing the limitations of current works by providing fine-grained preference feedbacks.


%% file: table/models.tex
\begin{table}
\setlength{\belowcaptionskip}{-0.02cm}
  \centering
\renewcommand\arraystretch{1}
\caption{An overview of popular text-to-video (T2V) and image-to-video (I2V) generation models. \textsuperscript{$\dag $} Representative variable.}
\vspace{-3mm}
   \resizebox{\linewidth}{!}{\begin{tabular}{lcccccc}
    \toprule[1pt]
     \bf Model &\bf Year&\bf Mode& \bf Resolution &\bf Frames &\bf Open\\
    \midrule
    CogVideo \cite{hong2022cogvideo}&22.05&T2V&480$\times$480&32&$\checkmark $\\
    Make-a-Video \cite{singer2022make}& 22.09 &T2V &256$\times$256&16&$\checkmark $\\
    LVDM~\cite{he2022latent}& 22.11 &T2V &256$\times$256&16&$\checkmark $\\
    Tune-A-Video~\cite{wu2023tune}& 22.12 &T2V &512$\times$512&8&$\checkmark $\\
    VideoFusion \cite{luo2023videofusion}& 23.03 &T2V &128$\times$128&16&$\checkmark $\\
    Text2Video-Zero \cite{khachatryan2023text2video}&23.03&T2V&512$\times$512&8&$\checkmark $\\
    ModelScope \cite{wang2023modelscope}&23.03&T2V&256$\times$256&16&$\checkmark $\\
    Lavie \cite{wang2023lavie}&23.09&T2V&512$\times$320&16&$\checkmark $\\
    VideoCrafter \cite{chen2023videocrafter1}
    &23.10&T2V, I2V&1024$\times$576&16&$\checkmark $\\
    Hotshot-XL \cite{Mullan_Hotshot-XL_2023}&23.10&T2V&672$\times$384&8&$\checkmark $\\
    StableVideoDiffusion \cite{blattmann2023stable} &23.11 & I2V &576$\times$1024&14&$\checkmark $\\
    AnimateDiff \cite{guo2023animatediff}&23.12&T2V, I2V&384$\times$256&20&$\checkmark $\\
    
    Floor33 \cite{floor33} & 23.08&T2V,I2V&1024$\times$640&16&$-$\\
    Genmo \cite{gem}&23.10&T2V, I2V&2048$\times$1536&60&$-$\\
    Gen-2 \cite{gen}&23.12&T2V, I2V&1408$\times$768&96&$-$\\
    MoonValley \cite{moonvalley}&24.01&T2V, I2V&1184$\times$672&200\textsuperscript{$\dag $}&$-$\\
    MorphStudio \cite{morph}&24.01&T2V, I2V&1920$\times$1080&72&$-$\\
    Sora~\cite{videoworldsimulators2024} & 24.02 &T2V, I2V&1920$\times$1080&600\textsuperscript{$\dag $}&$-$  \\
    
    \bottomrule[1pt]
  \end{tabular}}
  \label{Video_model}
 \vspace{-6mm}
\end{table}

%% file: table/database.tex
\begin{table*}[]
\vspace{-4mm}
\renewcommand\arraystretch{0.95}

  \caption{Summary of existing text-to-image and text-to-video evaluation datasets.}
  \vspace{-3mm}
  \tiny
  \label{tab:dataset}
    \centering
   \resizebox{\textwidth}{!}{\begin{tabular}{ccccccccc}
      \hline
    Dataset Types & Name & Numbers & Prompts & Models & Annotators & Dimensions & MOSs / Pairs & Annotation  \\
      \hline
    \multirow{5}{*}{AIGIQA}  
    &AGIQA-3k~\cite{li2023agiqa} & 2,982 & 180 & 6 & 21 & 2 & 5,964 & MOS \\
    &AIGCIQA2023~\cite{wang2023aigciqa2023} & 2,400 & 100 & 6 & 28 & 3 & 7,200 &MOS \\
    &RichHF-18k~\cite{liang2024rich}  & 17,760  & 17,760 & 3 &3 &4 &71,040 &MOS\\
    &HPS~\cite{wu2023better}& 98,807 & 25,205 & 1 & 2,659 & 1 & 25,205 &Pairs\\
    &Pick-a-Pic~\cite{kirstain2023pick} & - & 37,523 & 3 & 4,375 & 1 & 584,247&Pairs \\
    
     \hdashline
    \multirow{7}{*}{AIGVQA}&MQT~\cite{chivileva2023measuring} & 1,005 & 201 & 5 & 24 & 2 & 2,010 &MOS \\
    &EvalCrafter~\cite{liu2023evalcrafter} & 2,500 & 700 & 5 & 7 & 4 & 1,024 & MOS \\
    &FETV~\cite{liu2024fetv} & 2,476 & 619 & 4 & 3 & 3 & 7,428 & MOS\\
    &LGVQ~\cite{zhang2024benchmarking}  & 2,808 & 468 & 6 & 20& 3 & 8,424   &MOS\\
    &T2VQA-DB~\cite{kou2024subjective} & 10,000 & 1,000 & 9 & 27 & 1 & 10,000&MOS \\
    &GAIA~\cite{chen2024gaia} & 9,180 & 510 & 18 &54 &3 &27,540 &MOS\\
   \rowcolor{gray!20} &\textbf{ AIGVQA-DB(Ours)}& \textbf{36,576} & \textbf{1,048} & \textbf{15} &\textbf{120} &  \textbf{4}  &\textbf{122,304}& \textbf{MOS and Pairs}  \\
  \hline
  \end{tabular}}
    \vspace{-5mm}
\end{table*}

%% file: section/3_database.tex
\vspace{-2mm}
\section{Database Construction and Analysis}
\vspace{-1mm}
\subsection{Data Collection}
\input{figure/mosdis}
\input{figure/leida}
\input{figure/compare}
\vspace{-1mm}
\paragraph{Prompt Scources and Categorization.}
Prompts of the AIGVQA-DB are primarily sourced from existing open-domain text-video pair datasets, including InternVid \cite{wang2023internvid}, MSRVTT \cite{xu2016msr-vtt}, WebVid \cite{Bain21}, TGIF \cite{li2016tgif}, FETV \cite{liu2024fetv} and Sora website \cite{videoworldsimulators2024}.
We also manually craft prompts describing highly unusual scenarios to test the generalization ability of the generation models. 
As shown in Figure \ref{annotation}(a)-(d), we follow the categorization principles from FETV \cite{liu2024fetv} to organize each prompt based on the “spatial major content”,  “temporal major content”, “attribute control”, and “prompt complexity”.

\paragraph{Text-to-Video Generation.}
We utilize 15 latest text-to-video generative models to create AI-generated videos as shown in Figure \ref{annotation}(e). 
We leverage open-source website APIs and code with default weights for these models to produce AIGVs.
For the construction of the MOS subset, we collect 48 videos from the Sora Website \cite{videoworldsimulators2024}, along with their corresponding text prompts. Using these prompts, we generate additional videos using 11 different generative models. This process results in a total of 576 videos (12 generative models $\times$ 48 prompts). 
In addition to the MOS subset, we construct the pair-comparison subset using 1,000 diverse prompts, and 12 generative models including 8 open-sourced and 4 close-sourced are employed for text-to-video generation. Specifically, for each prompt, we generate four distinct videos for each open-source generative model and one video for each closed-source generative model. This process yields a total of 36,000 videos. More details of the database can be found in the \textit{Appendix} \ref{B}.

\vspace{-1mm}
\subsection{Subjective Experiment Setup and Procedure}
\vspace{-1mm}
Due to the unique and unnatural characteristics of AI-generated videos and the varying target video spaces dictated by different text prompts, relying solely on a single score, such as “quality”, to represent human visual preferences is insufficient.
In this paper, we propose to measure the human visual preferences of AIGVs from four perspectives.
\textbf{Static quality}  
assesses the clarity, sharpness, color accuracy, and overall aesthetic appeal of the frames when viewed as standalone images.
\textbf{Temporal smoothness} evaluates the temporal coherence of video frames and the absence of temporal artifacts such as flickering or jittering.
\textbf{Dynamic degree} 
evaluates the extent to which the video incorporates large motions and dynamic scenes, which contributes to the overall liveliness and engagement measurement of the content.
\textbf{Text-video (TV) correspondence} 
assesses how accurately the video content reflects the details, themes, and actions described in the prompt, ensuring that the generated video effectively translates the text input into a visual narrative.
Each of these four visual perception perspectives is related but distinct, offering a comprehensive evaluation for AIGVs. 
To evaluate the quality of the videos in the AIGVQA-DB, we conduct subjective experiments adhering to the guidelines outlined in ITU-R BT.500-14 \cite{duan2022confusing}.
For the MOS annotation type, we use a 1-5 Likert-scale judgment to score the videos. 
For the pairs annotation type, participants are presented with pairs of videos and asked to choose the one they prefer, providing a direct comparison method for evaluating relative video quality.
The videos are displayed using an interface designed with Python Tkinter, as illustrated in Figure \ref{annotation}(g)-(h).
A total of 120 graduate students participate in the experiment.

\vspace{-1mm}
\subsection{Subjective Data Processing}

\input{figure/model_score}
In order to obtain the MOS for an AIGV, we 
linearly scale the raw ratings to the range $[0,100]$ as follows:
$$z_i{}_j=\frac{r_i{}_j-\mu_i{}_j}{\sigma_i},\quad z_{ij}'=\frac{100(z_{ij}+3)}{6},$$
$$\mu_i=\frac{1}{N_i}\sum_{j=1}^{N_i}r_i{}_j, ~~ \sigma_i=\sqrt{\frac{1}{N_i-1}\sum_{j=1}^{N_i}{(r_i{}_j-\mu_i{}_j)^2}}$$ 
where $r_{ij}$ is the raw ratings given by the $i$-th subject to the $j$-th video. $N_i$ is the number of videos judged by subject $i$. 
Next, the MOS of the video j is computed by averaging the rescaled z-scores as follows:
$$MOS_j=\frac{1}{M}\sum_{i=1}^{M}z_{ij}'$$
where $MOS_j$ indicates the MOS for the $j$-th AIGV, $M$ is the number of subjects, and $z'_i{}_j$ are the rescaled z-scores. 

For the pairs annotation type, given a text prompt $p_{i}$, and 12 video generation models labeled $\{A, B, C, ... , L\}$, we generate videos using each model, forming a group of videos $G_{i,j}=\{V_{i,A,j}, V_{i,B,j}, V_{i,C,j}, ...,  V_{i,L,j}\}$. 
For each prompt $p_{i}$, we generate four different videos randomly for each of the eight open-source generative models and one video for each of the four closed-source generative models, resulting in a group of 36 videos $\{G_{i,A,1}, G_{i,A,2}, G_{i,A,3}, G_{i,A,4}, G_{i,B,1}, ...  G_{i,L,1}\}$.
For each group, we create all possible pairwise combinations, resulting in $C_{36}^2$ pairs: 
$(V_{A1}, V_{B1})$, $(V_{A1}, V_{B2})$, $(V_{A1}, V_{B3})$, $(V_{A1}, V_{B4})$, $(V_{A1}, V_{C1})$, ... , $(V_{K1}, V_{L1})$.
In the AIGVQA-DB construction pipeline, a prompt suite of 1000 prompts results in 630,000 ($1000 \times C_{36}^2$) pairwise video comparisons.
From this extensive dataset, we randomly sample 30,000 pairs for evaluation from four perspectives. Each pair is judged by three annotators, and the final decision of the better video in each pair is determined by the majority vote. 
Finally, we obtain a total of 46,080 reliable score ratings (20 annotators $\times$ 4 perspectives $\times$ 576 videos) and 360,000 pair ratings (3 annotators $\times$ 4 perspectives $\times$ 30,000 pairs).
\vspace{-1mm}
\subsection{AIGV Analysis from Four Perspectives}
As shown in Figure \ref{distribution}, the videos in the AIGVQA-DB cover a wide range of perceptual quality. 
We further analyze the win rates of various generation models across categories in Figure \ref{leida}, revealing the strengths and weaknesses of each T2V model. 
As shown in Figure \ref{leida}(a), the performances of T2V models rank uniform for different prompt complexity items in terms of static quality, which manifests current T2V model rank consistently for different prompts, likely due to shared architectures like diffusion-based systems, with common strengths and limitations in handling complex prompts. As shown in Figure \ref{leida}(b), in terms of attribute control, StableVideoDiffusion \cite{blattmann2023stable} excels in managing quantity over event order, as it first generates static images before animating them, preserving the original event sequence. As shown in Figure \ref{leida}(d), in terms of spatial content, most videos featuring “plants” and “people” show poor T2V correspondence. More comparison and analysis can be found in the \textit{Appendix} \ref{D}.
We also launch comparisons among text-to-video generation models regarding the MOS and pairwise win rates shown in Figure \ref{compare}. Notably, models such as LVDM~\cite{he2022latent} demonstrate exceptional performance in handling dynamic content, but exhibit relatively lower performance in temporal smoothness.
Sora~\cite{videoworldsimulators2024} and MorphStudio~\cite{morph} perform well in static quality and temporal smoothness while lagging in dynamic degree. Additionally, closed-source models exhibit much better performance compared to open-source models.

%% file: figure/mosdis.tex
\begin{figure*}[!t]
	\centering
	\includegraphics[width=\linewidth]{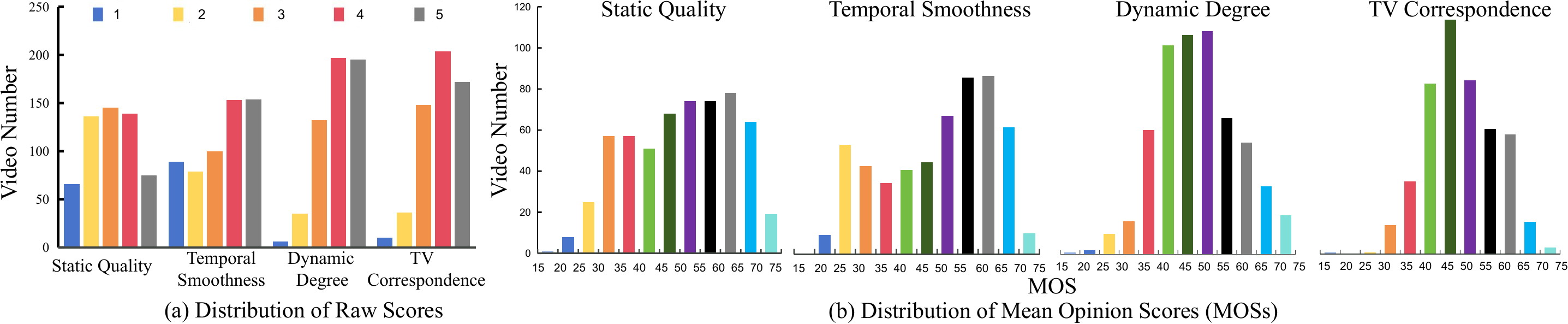}
 \vspace{-7mm}
	\caption{ Video score distribution from the four perspectives including static quality, temporal smoothness, dynamic degree, and t2v correspondence. (a) Distribution of raw scores. (b) Distribution of Mean Opinion Scores (MOSs)}
 \vspace{-1mm}
	\label{distribution}
\end{figure*}

%% file: figure/leida.tex
\begin{figure*}[!t]
	\centering
	\includegraphics[width=\linewidth]{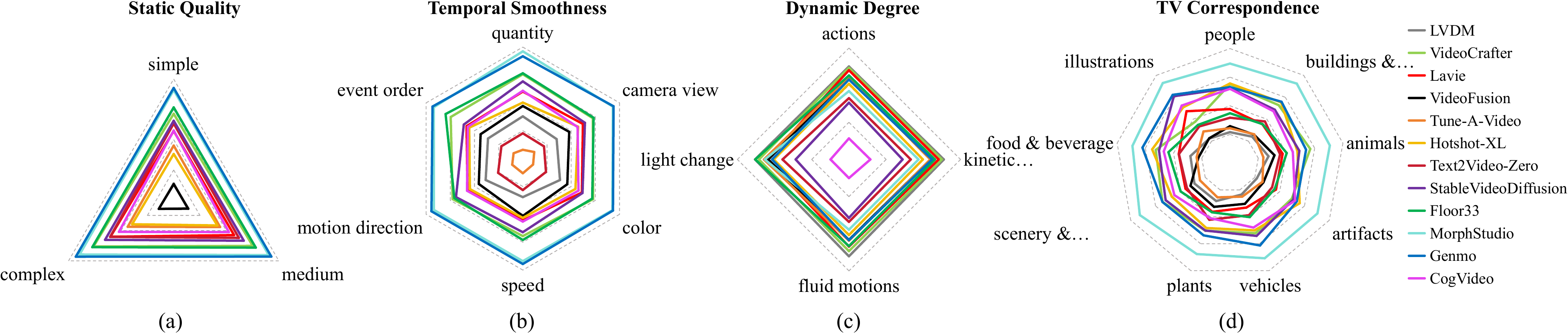}
 \vspace{-7mm}
	\caption{Comparison of averaged win rates of different generation models across different categories. (a) Results across prompt complexity. (b) Results across attribute control. (c) Results across temporal major contents. (d) Results across spatial major contents. }
 \vspace{-5mm}
	\label{leida}
\end{figure*}

%% file: figure/compare.tex
\begin{figure}[!t]
	\centering
	\includegraphics[width=\linewidth]{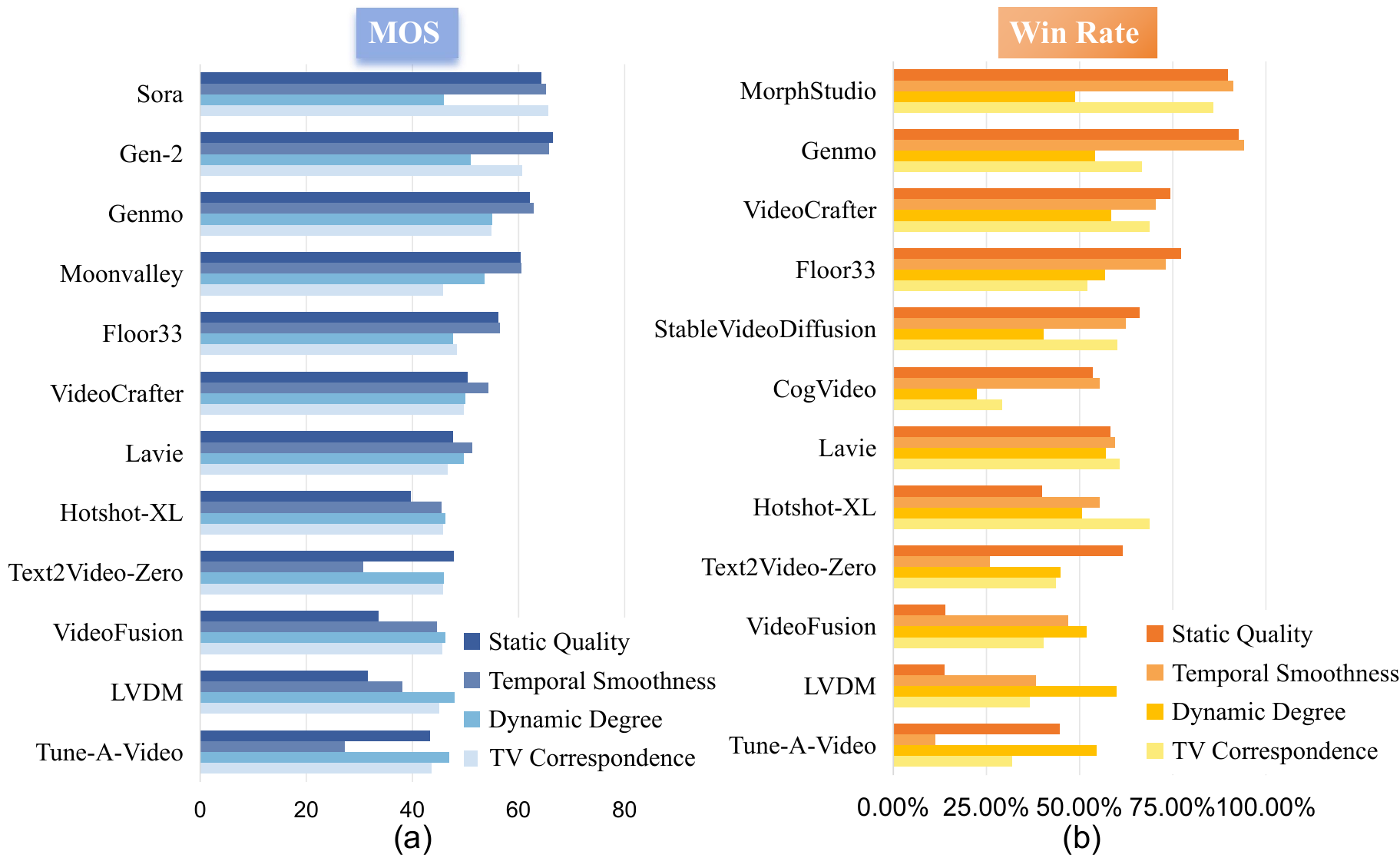}
 \vspace{-7mm}
	\caption{(a) Comparison of text-to-video generation models regarding the MOS in terms of four dimensions sorted bottom-up by their averaged MOS. (b) Comparison of text-to-video generation models regarding the win rate in terms of four dimensions sorted bottom-up by their averaged win rate. }
 \vspace{-4mm}
	\label{compare}
\end{figure}

%% file: figure/model_score.tex
\begin{figure*}[!t]
\vspace{-3mm}
	\centering
	\includegraphics[width=\linewidth]{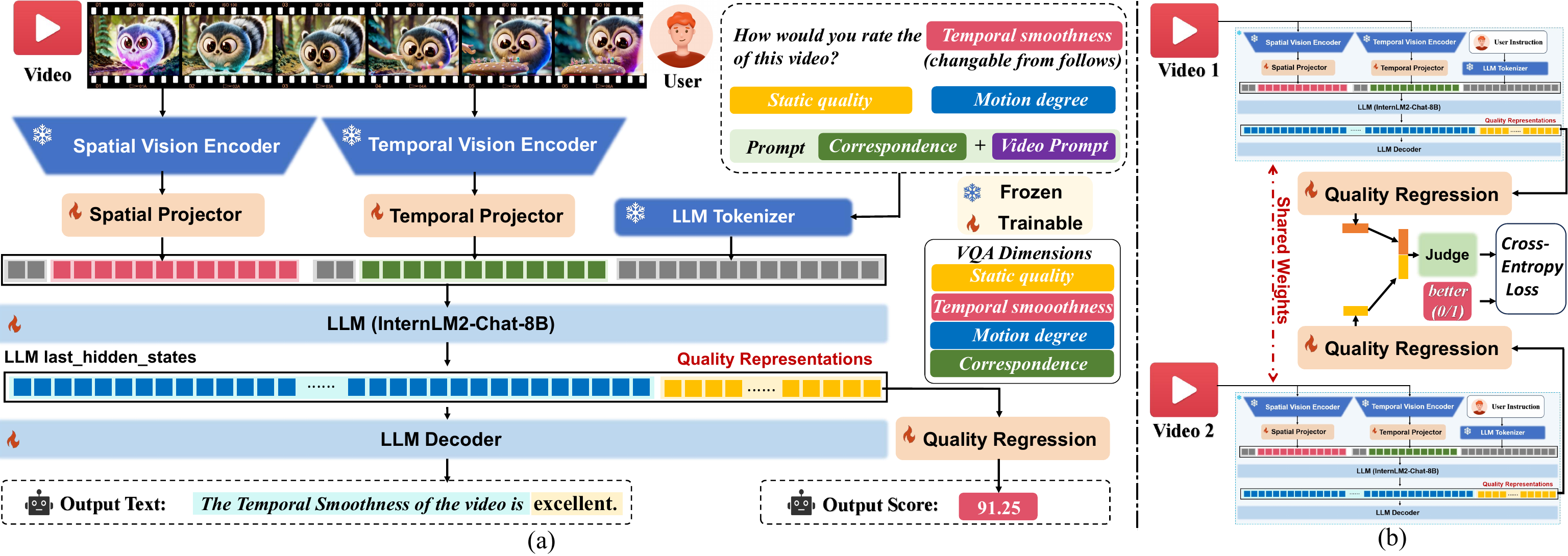}
 \vspace{-7mm}
	\caption{The framework of AIGV-Assessor: (a) AIGV-Assessor takes AI-generated video frames as input and outputs both text-based quality levels and numerical quality scores. The system begins with the extraction of spatiotemporal features using two vision encoders, which are then passed through spatial and temporal projection modules to generate aligned visual tokens into language space. The LLM decoder produces text-based feedback describing the video quality level for four evaluation dimensions, respectively. Simultaneously, the last-hidden-states from the LLM are used to perform quality regression that outputs final quality scores in terms of four dimensions. (b) AIGV-Assessor is fine-tuned on pairwise comparison, further allowing the model to output the evaluation comparison between two videos.   }
 \vspace{-3mm}
	\label{modelscore}
\end{figure*}

%% file: section/4_model.tex
\vspace{-1mm}
\section{Proposed Method}
\vspace{-1mm}
\subsection{Model Structure}
\vspace{-1mm}
\input{table/srcc}
\paragraph{Spatial and Temporal Vision Encoder.}
As shown in Figure \ref{modelscore}(a),
the model leverages two different types of encoders to capture the spatial and temporal characteristics of the video:
(1) 2D Encoder: A pre-trained 2D vision transformer (InternViT~\cite{wang2024charxiv}) is used to process individual video frames. 
(2) 3D Encoder: A 3D network, \textit{i.e.,} SlowFast~\cite{feichtenhofer2019slowfast}, is employed to extract temporal features by processing sequences of video frames. 

\paragraph{Spatiotemporal Projection Module.}
Once the spatial and temporal features are extracted, they are projected into a shared feature space for alignment with text-based queries. This is done through two projection modules that map the spatial and temporal visual features respectively into the language space. The mapped visual tokens are aligned with text tokens, enabling the model to query the video content in a multimodal fashion.

\paragraph{Feature Fusion and Quality Regression.}
We apply LLM (InternVL2-8B~\cite{wang2024charxiv}) to combine   the visual tokens and user-provided quality prompts to perform the following tasks:
(1) Quality level descriptions: the model generates a descriptive quality level evaluation of the input video, such as “The static quality of the video is (bad, poor, fair, good, excellent).”
This initial categorization provides a preliminary classification of the video's quality, which is beneficial for subsequent quality regression tasks. By obtaining a rough quality level, the model can more accurately predict numerical scores in later evaluations. 
(2) Regression score output: the model uses the final hidden states from the LLM to perform a regression task, outputting numerical quality scores for the video from four different dimensions.
\vspace{-1mm}
\subsection{Training and Fine-tuning Strategy}
\vspace{-1mm}
The training process of AIGV-Assessor follows a three-stage approach to ensure high-quality video assessment with quality level prediction, individual quality scoring, and pairwise preference comparison capabilities. This process includes: (1) training the spatial and temporal projectors to align visual and language features, (2) fine-tuning the vision encoder and LLM with LoRA \cite{hu2021lora}, and training the quality regression module to generate accurate quality scores, (3) incorporating pairwise comparison training using the pair-comparison subset with a pairwise loss function for robust video quality comparison.

\paragraph{Spatiotemporal Projector Training.}
The first stage focuses on training the spatial and temporal projectors to extract meaningful spatiotemporal visual features and map them into the language space. Through this process, the LLM is able to produce the quality level descriptions \textit{i.e.,} bad, poor, fair, good, excellent.

\paragraph{Quality Regression Fine-tuning.}
Once the model can generate coherent descriptions of video quality level, the second stage focuses on fine-tuning the quality regression module. The goal here is to enable the model to output stable and precise numerical quality scores (MOS-like predictions).
The quality regression model takes the last-hidden-state features from LLM as input and generates quality scores from four perspectives. 
The training objective uses an L1 loss function to minimize the difference between the predicted quality score and the groundtruth MOS.

\input{table/lgvq}

 \paragraph{Pairwise Comparison Fine-tuning.}
 The third stage mainly focuses on integrating the pairwise comparison into the training pipeline.
 As shown in Figure \ref{modelscore}(b), two input video pairs share network weights within the same batch. We design a judge network inspired by LPIPS \cite{zhang2018unreasonable} to determine which video performs better. 
 This network leverages learned features and evaluates the perceptual differences between the two videos, allowing more reliable quality assessments in video pair comparison.

\paragraph{Loss Function.}
 In the first stage, the spatial and temporal projectors are trained to align visual and language features using language loss. The second stage refines the vision encoder, LLM, and  quality regression module’s scoring ability with an L1 loss. The third stage incorporates pairwise comparison training with cross-entropy loss to improve the model's performance on relative quality evaluation.

%% file: table/srcc.tex
\begin{table*}[tbph]
\vspace{-4mm}
\renewcommand\arraystretch{1.15}
\caption{Performance comparisons of the state-of-the-art quality evaluation methods on the AIGVQA-DB from four perspectives. The best performance results are marked in {\textcolor{red}{RED}} and the second-best performance results are marked in {\blue{BLUE}}.}
  \label{t2}
  \centering
\vspace{-3mm}

\resizebox{\textwidth}{!}{
  \begin{tabular}{l||c:ccc|c:ccc|c:ccc|c:ccc}
    \hline
    \multicolumn{1}{l}{\bf Dimension} &
    \multicolumn{4}{c}{\textbf{Static Quality}} &
    \multicolumn{4}{c}{\textbf{Temporal Smoothness}} &
    \multicolumn{4}{c}{\textbf{Dynamic Degree}} &
    \multicolumn{4}{c}{\textbf{TV Correspondence}} 
    \\
    \cmidrule(lr){2-5}\cmidrule(lr){6-9}\cmidrule(lr){10-13}\cmidrule(lr){14-17}
\textbf{Methods / Metrics}  
& \textbf{Pair Acc} & \textbf{SRCC}  & \textbf{PLCC}  & \textbf{KRCC}   & \textbf{Pair Acc}  & \textbf{SRCC}  & \textbf{PLCC}  & \textbf{KRCC}  & \textbf{Pair Acc}  & \textbf{SRCC}  & \textbf{PLCC}  & \textbf{KRCC}   & \textbf{Pair Acc} & \textbf{SRCC}  & \textbf{PLCC}  & \textbf{KRCC}    \\

\hline
\textbf{NIQE} \cite{mittal2012making}
& 54.32\% & 0.0867 & 0.1626 & 0.0615 
& 52.67\% & 0.0641 & 0.1152 & 0.0451  
& 45.64\% & 0.1765 & 0.2448 & 0.1194 
& 46.99\% & 0.1771 & 0.2231 & 0.1193 
\\
\textbf{QAC} \cite{xue2013learning} 
& 49.96\% & 0.1022  & 0.1363 & 0.0680 
& 54.90\% & 0.1633  & 0.2039 & 0.1105 
& 54.72\% & 0.0448  & 0.0427 & 0.0295 
& 54.48\% & 0.0303  & 0.0197 & 0.2233 
\\
\textbf{BRISQUE} \cite{mittal2012no} 
& 59.98\% & 0.2909 & 0.2443 & 0.1969 
& 55.67\% & 0.2325 & 0.1569 & 0.1553 
& 44.60\% & 0.1351 & 0.0959 & 0.0893 
& 51.02\% & 0.1294 & 0.1017 & 0.0869 
\\
\textbf{BPRI}  \cite{min2017blind}
& 52.28\% & 0.2181 & 0.1723 & 0.1398 
& 47.26\% & 0.1766 & 0.0880 & 0.1138 
& 46.83\% & 0.1956 & 0.1688 & 0.1329 
& 49.13\% & 0.1569 & 0.1548 & 0.1052 
\\
 
\textbf{HOSA} \cite{xu2016blind} 
& 61.54\% & 0.2420 & 0.2106 & 0.1643 
& 57.31\% & 0.2311 & 0.1757 & 0.1559 
& 44.97\% & 0.0755 & 0.0449 & 0.0496 
& 52.23\% & 0.1645  &0.1324 & 0.1097 
\\
\textbf{BMPRI} \cite{quality:BMPRI} 
& 53.71\% & 0.1690 & 0.1481 & 0.1075 
& 49.31\% & 0.1434 & 0.0844 & 0.0894 
& 45.07\% & 0.1153 & 0.0925 & 0.0777 
& 48.43\% & 0.1567 & 0.1500 & 0.1041 
\\
\textbf{V-Dynamic}~\cite{huang2024vbench}
& 51.34\% & 0.0768 & 0.0792 & 0.0494 
& 31.91\% & 0.3713 & 0.4871 & 0.2557 
& 53.11\% & 0.1466 & 0.0253 & 0.0988 
& 46.96\% & 0.0405 & 0.0576 & 0.0223 
\\
\textbf{V-Smoothness}~\cite{huang2024vbench}
& 61.63\% & 0.6748 & 0.4506 & 0.4590 
& 76.59\% & 0.8526 & 0.8313 & 0.6533 
& 47.63\% & 0.2446 & 0.2328 & 0.1580 
& 61.28\% & 0.3188 & 0.3073 & 0.2214 
\\

\midrule



\textbf{CLIPScore} \cite{CLIPScore}
& 47.09\% & 0.0731 & 0.0816 & 0.0473 
& 46.33\% & 0.0423 & 0.0334 & 0.0271 
& 52.99\% & 0.0675 & 0.0835 & 0.0439 
& 55.62\% & 0.1519 & 0.1731 & 0.1014 
\\
\textbf{BLIPScore} \cite{li2022blip}
& 53.24\% & 0.0492 & 0.0421 & 0.0330 
& 53.07\% & 0.0659 & 0.0487 & 0.0437 
& 53.03\% & 0.1786 & 0.1904 & 0.1205 
& 61.53\% & 0.1813 & 0.1896 & 0.1219 
\\
\textbf{AestheticScore} \cite{schuhmann2022laion}
& 70.24\% & 0.6713 & 0.6959 & 0.4784 
& 54.82\% & 0.5154 & 0.4946 & 0.3484 
& 52.96\% & 0.2295 & 0.2322 & 0.1527 
& 59.64\% & 0.2381 & 0.2440 & 0.1602 
\\
\textbf{ImageReward} \cite{database/align:ImageReward}
& 56.69\% & 0.2606 & 0.2646 & 0.1749 
& 54.09\% & 0.2382 & 0.2305 & 0.1600 
& 53.90\% & 0.1840 & 0.1836 & 0.1237 
& 63.97\% & 0.2311 & 0.2450 & 0.1568 
\\
\textbf{UMTScore} \cite{liu2024fetv}
& 48.93\% & 0.0168 & 0.0199 & 0.0117 
& 49.93\% & 0.0302 & 0.0370 & 0.0207 
& 52.69\% & 0.0168 & 0.0198 & 0.0117 
& 53.82\% & 0.0172 & 0.0065 & 0.0108 

\\
\midrule

\textbf{Video-LLaVA} \cite{lin2023video}
& 50.90\% & 0.0384 & 0.0513 & 0.0297
& 50.36\% & 0.0431 & 0.0281 & 0.0347
& 50.34\% & 0.1561 & 0.1436 & 0.1176
& 50.54\% & 0.1364 & 0.1051 & 0.1009
\\
\textbf{Video-ChatGPT} \cite{Maaz2023VideoChatGPT}
& 51.20\% & 0.1242 & 0.1587 & 0.0940
& 50.16\% & 0.0580 & 0.0533 & 0.0453 
& 50.47\% & 0.0724 & 0.0436 & 0.0563
& 50.07\% & 0.0357 & 0.0124 & 0.0274
\\
\textbf{LLaVA-NeXT} \cite{li2024llava}
& 52.85\% & 0.1239 & 0.1625 & 0.0954
& 52.41\% & 0.4021 & 0.3722 & 0.3052
& 51.84\% & 0.1767 & 0.1655 & 0.1328 
& 59.20\% & 0.4116 & 0.3428 & 0.3261 
\\
\textbf{VideoLLaMA2} \cite{damonlpsg2024videollama2}
& 52.73\% & 0.2643 & 0.3271 & 0.1928
& 52.27\% & 0.3608 & 0.2450 & 0.2696
& 50.78\% & 0.1900 & 0.1561 & 0.1379 
& 54.25\% & 0.1656 & 0.1633 & 0.1210 
\\
\textbf{Qwen2-VL} \cite{Qwen2VL}
& 56.50\% & 0.4922 & 0.5291 & 0.3838 
& 49.12\% & 0.1681 & 0.4219 & 0.1233
& 52.08\% & 0.1122 & 0.1335 & 0.0849
& 53.30\% & 0.3111 & 0.2775 & 0.2306
\\
\midrule
\textbf{HyperIQA}~\cite{Su_2020_CVPR}
& 68.30\% & 0.7931 & 0.8093 & 0.5969 
& 54.65\% & 0.7426 & 0.6630 & 0.5407
& 53.32\% & 0.2103 & 0.2100 & 0.1384 
& 57.54\% & 0.6226 & 0.6250 & 0.4432 
\\
\textbf{MUSIQ}~\cite{MUSIQ}
& 66.46\% & 0.7880 & 0.8044 & 0.5773 
& 55.16\% & 0.7199 & 0.6920 & 0.5034 
& 52.85\% & 0.5206 & 0.4846 & 0.3521 
& 58.46\% & 0.4125 & 0.4093 & 0.2844 
\\
\textbf{LIQE}~\cite{LIQE}
& 63.86\% &0.8776 & 0.8691 & 0.7008 
& 55.84\% &0.7935 & 0.7720 & 0.6084 
& 49.02\% &0.5303 & 0.5840 & 0.3837 
& 55.10\% &0.3862 & 0.3639 & 0.2640 
\\
\textbf{VSFA}~\cite{VSFA}
& 46.43\% & 0.3365 & 0.3421 & 0.2268 
& 50.95\% & 0.3317 & 0.3273 & 0.2202 
& 51.46\% & 0.1201 & 0.1362 & 0.0815 
& 48.07\% & 0.1024 & 0.1064 & 0.0666 
\\
 \textbf{BVQA}~\cite{li2022blindly} 
& 29.98\% & 0.4594 & 0.4701 & 0.3268 
& 37.65\% & 0.3704 & 0.3819 & 0.2507 
& \bf\blue{55.08\%} & 0.4594 & 0.4701 & 0.3268 
& 42.32\% & 0.3720 & 0.3978 & 0.2559 
\\
\textbf{simpleVQA}~\cite{sun2022a} 
& 68.12\% & 0.8355 & 0.6438 & 0.8489 
& 54.14\% & 0.7082 & 0.7008 & 0.4978 
& 53.08\% & 0.4671 & 0.3160 & \bf\blue{0.3994} 
& 58.20\% & 0.4643 & 0.5440 & 0.3163 
\\
\textbf{FAST-VQA}~\cite{wu2022fast}
& 70.64\% & 0.8738 & 0.8644 & 0.6860 
& \bf\blue{62.93\%} & 0.9036 & 0.9134 & 0.7166 
& 54.34\% & 0.5603 & \bf\blue{0.5703} & 0.3895 
& \bf\blue{65.05\%}
& \bf\blue{0.6875} & \bf\blue{0.6704}  
& \bf\blue{0.4978}
\\
\textbf{DOVER}~\cite{wu2023dover} 
&  \bf\blue{72.92\%}
& \bf\blue{0.8907} & \bf\blue{0.8895} 
& \bf\blue{0.7004} 
& 58.83\% & \bf\blue{0.9063} & \bf\blue{0.9195} 
& \bf\blue{0.7187} 
& 53.16\% & 0.5549 & 0.5489  & 0.3800 
& 62.35\% & 0.6783 & 0.6802  & 0.4969 
\\
\textbf{Q-Align} \cite{wu2023q}
& 71.86\% & 0.8516 & 0.8383 & 0.6641 
& 57.95\% & 0.8116 & 0.7025 & 0.6195 
& 53.71\% & \bf\blue{0.5655} & 0.5012 
& 0.3950 
& 62.91\% & 0.5542 & 0.5647 & 0.3870 
\\
\hline
\rowcolor{gray!20}\textbf{Ours} 
&\bf\textcolor{red}{79.83\%}
&\bf\textcolor{red}{0.9162} &\bf\textcolor{red}{0.9190}
&\bf\textcolor{red}{0.7576} 
&\bf\textcolor{red}{76.60\%}
&\bf\textcolor{red}{0.9232} &\bf\textcolor{red}{0.9216}  
&\bf\textcolor{red}{0.8038} 
&\bf\textcolor{red}{60.30\%}
&\bf\textcolor{red}{0.6093} &\bf\textcolor{red}{0.6082}  
&\bf\textcolor{red}{0.4435}  
&\bf\textcolor{red}{70.32\%}
&\bf\textcolor{red}{0.7500} &\bf\textcolor{red}{0.7697}  
&\bf\textcolor{red}{0.5591} 
\\
\rowcolor{gray!20}\textit{Improvement} 
& + 6.9\%
& +2.7\% & +3.0\% &+ 5.7\%  
& 13.7\%
& +1.7\% & +0.2\% &+8.5\%  
& +5.2\% 
& +4.4\% & +3.8\% &+ 4.4\% 
& +5.3\%
& +6.3\% & +9.9\% &+6.13\% \\
\hline
\end{tabular}}\label{mos}
 \vspace{-4mm}
\end{table*}

%% file: table/lgvq.tex
\begin{table}[]
\vspace{-4mm}
\caption{Performance comparisons on LGVQ~\cite{zhang2024benchmarking} and FETV~\cite{liu2024fetv}.}
\label{lgvq}
\vspace{-3mm}
\renewcommand\arraystretch{1.1}
\resizebox{0.48\textwidth}{!}{
\begin{tabular}{clllllll}
\toprule
\multirow{2}{*}{Aspects}   
& \multicolumn{1}{c}{\multirow{2}{*}{Methods}} 
& \multicolumn{3}{c}{ LGVQ}                                        
& \multicolumn{3}{c}{ FETV}                                         
  
\\
\cmidrule(r){3-5} \cmidrule(r){6-8} 
& \multicolumn{1}{c}{} & SRCC & PLCC & KRCC & SRCC & PLCC & KRCC\\
\midrule  
\multirow{6}{*}{Spatial}   
& MUSIQ~\cite{MUSIQ}   
& 0.669  & 0.682 & 0.491 & 0.722  & 0.758 & 0.613\\
& StairIQA~\cite{StairIQA} 
& 0.701  & 0.737 & 0.521 & 0.806  & 0.812 & 0.643\\
& CLIP-IQA~\cite{CLIP_IQA} 
& 0.684  & 0.709 & 0.502 & 0.741  & 0.767 & 0.619\\
& LIQE~\cite{LIQE} 
& 0.721  & 0.752 & 0.538 & 0.765  & 0.799 & 0.635\\
& UGVQ~\cite{zhang2024benchmarking} 
& \bf\blue{0.759}  & \bf\blue{0.795} & \bf\blue{0.567} & \bf\blue{0.841}  & \bf\blue{0.841} & \bf\blue{0.685}\\
\cdashline{2-8}
\rowcolor{gray!20}& \textbf{Ours} & \bf\textcolor{red}{0.803} & \bf\textcolor{red}{0.819}  & \bf\textcolor{red}{0.617} 
                & \bf\textcolor{red}{0.853}  & \bf\textcolor{red}{0.856} & \bf\textcolor{red}{0.699}\\
\rowcolor{gray!20}& \textit{Improvement} & + 4.4\% & +2.4\% & +5.0\% &+1.2\% & +1.5\% & +1.4\%\\
\midrule
\multirow{7}{*}{Temporal}  
& VSFA~\cite{VSFA}      
& 0.841  & 0.857 & 0.643  & 0.839 & 0.859 & 0.705  \\
& SimpleVQA~\cite{sun2022a} 
& 0.857 & 0.867  & 0.659  & 0.852 & 0.862 & 0.726  \\
& FastVQA~\cite{wu2022fast}
& 0.849 & 0.843 & 0.647  & 0.842 & 0.847 & 0.714  \\
& DOVER~\cite{wu2023dover} 
& 0.867 & 0.878 & 0.672  & 0.868 & 0.881 & 0.731  \\
& UGVQ~\cite{zhang2024benchmarking}      & \bf\blue{0.893} & \bf\blue{0.907} & \bf\blue{0.703}  & \bf\blue{0.897} & \bf\blue{0.907} & \bf\blue{0.753}  \\
\cdashline{2-8}
\rowcolor{gray!20}& \textbf{Ours}  &\bf\textcolor{red}{0.900} & \bf\textcolor{red}{0.920} & \bf\textcolor{red}{0.717}   
& \bf\textcolor{red}{0.936} &\bf\textcolor{red}{0.940} & \bf\textcolor{red}{0.815}  \\
\rowcolor{gray!20}& \textit{Improvement} & +0.7\% & +1.3\% &+1.4\% & +3.9\% & +3.3\% & +6.2\%\\
\midrule
\multirow{8}{*}{Alignment} 
& CLIPScore~\cite{CLIPScore}   
& 0.446 & 0.453 & 0.301  & 0.607 & 0.633 & 0.498  \\
& BLIPScore~\cite{li2022blip} 
& 0.455 & 0.464 & 0.319  & 0.616 & 0.645 & 0.505  \\
& ImageReward~\cite{database/align:ImageReward}
& 0.498 & 0.499 & 0.344  & 0.657 & 0.687 & 0.519  \\
& PickScore~\cite{PickScore}  
& 0.501 & 0.515 & 0.353  & 0.669 & 0.708 & 0.533  \\
& HPSv2~\cite{HPSv2}      
& 0.504 & 0.511 & 0.357  & 0.686 & 0.703 & 0.540  \\
& UGVQ~\cite{zhang2024benchmarking}        
& \bf\blue{0.551} & \bf\blue{0.555} & \bf\blue{0.394}  & \bf\blue{0.734} & \bf\blue{0.737} & \bf\blue{0.572}  \\
\cdashline{2-8}
\rowcolor{gray!20} & \textbf{Ours} & \bf\textcolor{red}{0.577}  & \bf\textcolor{red}{0.578} & \bf\textcolor{red}{0.411}
& \bf\textcolor{red}{0.753}  & \bf\textcolor{red}{0.746} & \bf\textcolor{red}{0.585}\\
\rowcolor{gray!20}&\textit{Improvement} & +2.6\% &+2.3\% & +1.7\% & +1.9\% &+0.9\% &+1.3\%\\
\bottomrule
\end{tabular}}
\vspace{-5mm}
\end{table}

%% file: section/5_exp.tex
\vspace{-1mm}
\section{Experiments}
\vspace{-1mm}
\subsection{Experiment Settings}
\vspace{-1mm}
\paragraph{Evaluation Datasets and Metrics.}
Our proposed method is validated on five AIGVQA datasets: AIGVQA-DB, LGVQ~\cite{zhang2024benchmarking}, FETV~\cite{liu2024fetv}, T2VQA~\cite{kou2024subjective}, and GAIA~\cite{chen2024gaia}. To evaluate the correlation between the predicted scores and the ground-truth MOSs, we utilize three evaluation criteria: Spearman Rank Correlation Coefficient (SRCC), Pearson Linear Correlation Coefficient (PLCC), and Kendall’s Rank Correlation Coefficient (KRCC).  For pair comparison, we adopt the comparison accuracy as the metric.

\paragraph{Reference Algorithms.}
To assess the performance of our proposed method, we select state-of-the-art evaluation metrics for comparison, which can be classified into five groups:
(1) Handcrafted-based I/VQA models, including: NIQE \cite{mittal2012making}, BRISQUE \cite{mittal2012no}, QAC \cite{xue2013learning}, BMPRI \cite{quality:BMPRI}, HOSA \cite{xu2016blind}, BPRI \cite{min2017blind},    HIGRADE \cite{kundu2017large}, \textit{etc.} 
(2) Action-related evaluation models, including: V-Dynamic~\cite{huang2024vbench}, V-Smoothness~\cite{huang2024vbench}  which are proposed in VBench~\cite{huang2024vbench}.
(3) Vision-language pre-training models, including: CLIPScore~\cite{CLIPScore}, BLIPScore \cite{li2022blip}, AestheticScore \cite{schuhmann2022laion},  ImageReward \cite{database/align:ImageReward}, and UMTScore \cite{liu2024fetv}.
(4) LLM-based models, including: Video-LLaVA \cite{lin2023video}, Video-ChatGPT \cite{Maaz2023VideoChatGPT}, LLaVA-NeXT \cite{li2024llava}, VideoLLaMA2 \cite{damonlpsg2024videollama2}, and Qwen2-VL \cite{Qwen2VL}.
(5) Deep learning-based I/VQA models, including: HyperIQA~\cite{Su_2020_CVPR}, MUSIQ~\cite{MUSIQ}, LIQE~\cite{LIQE}, VSFA \cite{VSFA}, BVQA \cite{li2022blindly}, SimpleVQA \cite{sun2022deep}, FAST-VQA \cite{wu2022fast}, DOVER \cite{wu2023dover}, and Q-Align \cite{wu2023q}.
\input{table/t2vqa}
\input{table/gaia}

\paragraph{Training Settings.}
Traditional handcrafted models are directly evaluated on the corresponding databases, and the average score of all frames is calculated.
For vision-language pre-training and LLM-based models, we load the pre-trained weights for inference. 
CLIPscore~\cite{CLIPScore}, BLIPscore \cite{li2022blip}, and other vision-language pre-training models are calculated directly as the average cosine similarity between text and each video frame. 
SimpleVQA \cite{sun2022deep}, BVQA \cite{li2022blindly}, FAST-VQA \cite{wu2022fast}, DOVER \cite{wu2023dover}, and Q-Align \cite{wu2023q} are fine-tuned on every test dataset. 
For deep learning-based IQA and VQA models, all experiments for each method are retrained on each dimension using the same training and testing split as the previous literature at a ratio of 4:1.
All results are averaged after ten random splits.
\vspace{-1mm}
\subsection{Results and Analysis}
\vspace{-1mm}
 \input{figure/result}
\input{table/ablation}
Table \ref{mos} presents the pairwise win rates and the score prediction correlation between predicted results and human ground truths. The results indicate that handcrafted-based methods consistently underperform across all four evaluation perspectives.
Vision-language pre-training methods such as CLIPscore~\cite{CLIPScore} and BLIPscore \cite{li2022blip} demonstrate moderate performance but are still surpassed by more specialized and fine-tuned VQA models. 
 Specifically, deep learning-based models like FAST-VQA \cite{wu2022fast} and DOVER \cite{wu2023dover} achieve more competitive performances after fine-tuning.
 However, they are still far away from satisfactory.
 Notably, most VQA models perform better on quality evaluation than on text-video correspondence, as they lack text prompts input used in video generation, making it challenging to extract relation features from the AI-generated videos, which inevitably leads to the performance drop.
Finally, the performance exploration of recent LMMs on our database shows that current LMMs are able to produce meaningful evaluations, which can motivate future works to further explore the use of LMMs for AIGV assessment.

The proposed AIGV-Assessor achieves the best performance compared to the competitors for both MOS prediction and pair ranking tasks in terms of all four dimensions.
To further validate the effectiveness and generalizability of our proposed model, we also evaluate it on four other AIGVQA datasets \cite{zhang2024benchmarking,liu2024fetv,kou2024subjective,chen2024gaia}. From Tables \ref{lgvq}-\ref{gaia}, we observe that AIGV-Assessor consistently achieves the best performance across these datasets. 
As shown in Figure \ref{leida2}, AIGV-Assessor achieves the highest overlap in area with Ground Truth (GT),
indicating that AIGV-Assessor can reliably perform T2V model benchmarking, outperforming other assessment models in discerning quality differences in AI-generated videos.

\vspace{-1mm}
\subsection{Ablation Study}
\vspace{-1mm}
We conduct ablation experiments to verify the effectiveness of the main components in our AIGV-Assessor method, including the spatial feature, the temporal feature, the quality level, and the LoRA finetuning strategy.
Additionally, we assess how each feature contributes to the performance across different quality dimensions. The results of these experiments are summarized in Table \ref{ablation}. 
Experiments (1), (2), and (3) validate the effectiveness of the quality regression module and the LoRA finetuning strategy, confirming that fine-tuning and quality regression significantly enhance model performance over only regressing the generated text outputs from the LLM.
 The addition of temporal features, as seen in Experiments (4), (5), and (6), significantly improves model performance.
 Experiment (6), which integrates all components, yields the best overall performance, showing that the combination of spatial and temporal features, quality level prediction, and LoRA finetuning provides the most robust and accurate AIGV assessment.
 \vspace{-2mm}

%% file: table/t2vqa.tex
\begin{table}[]
\vspace{-4mm}
\caption{Performance comparisons on T2VQA-DB~\cite{kou2024subjective}.}
\vspace{-3mm}
\label{t2vqa}
\renewcommand\arraystretch{1.1}
\resizebox{0.48\textwidth}{!}{
\begin{tabular}{clllllll}
\toprule
\multirow{2}{*}{Aspects}   
& \multicolumn{1}{c}{\multirow{2}{*}{Methods}} 
& \multicolumn{3}{c}{T2VQA-DB}                                        
& \multicolumn{3}{c}{Sora Testing}                                         
  
\\
\cmidrule(r){3-5} \cmidrule(r){6-8} 
& \multicolumn{1}{c}{} & SRCC & PLCC & KRCC & SRCC & PLCC & KRCC\\
\midrule  
\multirow{4}{*}{zero-shot}   
& CLIPScore ~\cite{CLIPScore}
& 0.1047  & 0.1277 & 0.0702 & 0.2116  & 0.1538 & 0.1406\\
& BLIPScore~\cite{li2022blip}
& 0.1659  & 0.1860 & 0.1112 & 0.2116  & 0.1038 & 0.1515\\
& ImageReward~\cite{database/align:ImageReward}
& 0.1875  & 0.2121 & 0.1266 & 0.0992  & 0.0415 & 0.0748\\
& UMTScore~\cite{liu2024fetv}
& 0.0676  & 0.0721 & 0.0453 & 0.2594  & 0.0840 & 0.1680\\
\midrule
\multirow{5}{*}{finetuned}   
& SimpleVQA~\cite{sun2022a} 
& 0.6275  & 0.6388 & 0.4466 & 0.0340  & 0.2344 & 0.0237\\
& BVQA~\cite{li2022blip} 
& 0.7390  & 0.7486 & 0.5487 & 0.4235  & 0.2489 & 0.2635\\
& FAST-VQA~\cite{wu2022fast}
& 0.7173  & 0.7295 & 0.5303 & 0.4301  & 0.2369 & 0.2939\\
& DOVER~\cite{wu2023dover} 
& 0.7609 & 0.7693 & 0.5704 & 0.4421  & 0.2689 & 0.2757\\
& T2VQA~\cite{kou2024subjective} &\bf\blue{0.7965} & \bf\blue{0.8066} &\bf\blue{0.6058} & \bf\blue{0.6485} &\bf\blue{0.3124} & \bf\blue{0.4874}\\
\midrule
\rowcolor{gray!20}& \textbf{Ours} & \bf\textcolor{red}{0.8131} & \bf\textcolor{red}{0.8222}  &\bf\textcolor{red}{0.6364} 
                & \bf\textcolor{red}{0.6612}  &\bf\textcolor{red}{0.3318} & \bf\textcolor{red}{0.5075}\\
\rowcolor{gray!20}& \textit{Improvement} & + 1.7\% & +1.6\% &+3.1\% &+1.3\%& +1.9\% & +2.0\%\\
\bottomrule
\end{tabular}}
\vspace{-2mm}
\end{table}

%% file: table/gaia.tex
\begin{table}
\setlength{\belowcaptionskip}{-0.01cm}
\centering
\belowrulesep=0pt
\aboverulesep=0pt
\renewcommand\arraystretch{1.2}
\caption{ Performance comparisons on GAIA~\cite{chen2024gaia}. 
}
\vspace{-3mm}
   \resizebox{\linewidth}{!}{\begin{tabular}{lcccccc}
    \toprule
    Dimension  &\multicolumn{2}{c}{Subject}&\multicolumn{2}{c}{Completeness}&\multicolumn{2}{c}{Interaction}\\
  \cmidrule(lr){2-3} \cmidrule(lr){4-5} \cmidrule(lr){6-7} 
 Methods / Metrics
 &SRCC&PLCC&SRCC&PLCC&SRCC&PLCC\\
    \midrule
  V-Smoothness~\cite{huang2024vbench}
    &0.2402&0.1913&0.1474&0.1625&0.1741&0.1693\\
   V-Dynamic~\cite{huang2024vbench}
    &0.1285&0.0831&0.0903&0.0682&0.1141&0.0758\\
    Action-Score~\cite{liu2023evalcrafter}
    &0.2023&0.1823&0.2867&0.2623&0.2689&0.2432 \\
  Flow-Score \cite{liu2023evalcrafter}
   &0.1471&0.1541&0.0816&0.1273&0.1041&0.1309\\
    CLIPScore~\cite{CLIPScore}
    &0.3398&0.3330&0.3944&0.3871&0.3875&0.3821 \\
    BLIPScore~\cite{li2022blip}
    &0.3453&0.3386&0.4174&0.4082&0.4044&0.3994 \\
    LLaVAScore~\cite{liu2024visual}
    &0.3484&0.3436&0.4189&0.4133&0.4077&0.4025\\
    \hdashline
    TLVQM~\cite{korhonen2019two}
    &0.5037&0.5137&0.4127&0.4158&0.4079&0.4093\\
    VIDEVAL~\cite{tu2021ugc}
    &0.5237&0.5446&0.4283&0.4375&0.4121&0.4234\\
    VSFA~\cite{VSFA}
    &0.5594&0.5762&0.4940&0.5017&0.4709&0.4811\\
    BVQA~\cite{li2022blip} 
    &0.5702&0.5888&0.4876&0.4946&0.4761&0.4825\\
    SimpleVQA~\cite{sun2022deep}
    &0.5920&0.5974&0.4981&0.5078&0.4843&0.4971\\
    FAST-VQA~\cite{wu2022fast}
    &0.6015&0.6092&0.5157&0.5215&0.5154&0.5216\\ 
    DOVER~\cite{wu2023dover} 
   & \bf\blue{0.6173}& \bf\blue{0.6301}& \bf\blue{0.5198}
   & \bf\blue{0.5323}& \bf\blue{0.5164}& \bf\blue{0.5278}\\

   \hdashline
    \rowcolor{gray!20} \textbf{Ours} &\bf\textcolor{red}{ 0.6842} & \bf\textcolor{red}{0.6897} & \bf\textcolor{red}{0.6635 }& \bf\textcolor{red}{0.6694} & \bf\textcolor{red}{0.6329 }& \bf\textcolor{red}{0.6340}\\
   \rowcolor{gray!20}\textit{Improvement} & +6.7\% & +6.0\% & +14.4\% & +13.7\% & +11.65\% &+10.6\%\\
    \bottomrule
    
  \end{tabular}}
  \label{gaia}
  \vspace{-5mm}
\end{table}

%% file: figure/result.tex
\begin{figure*}[!t]
\vspace{-5mm}
	\centering
	\includegraphics[width=\linewidth]{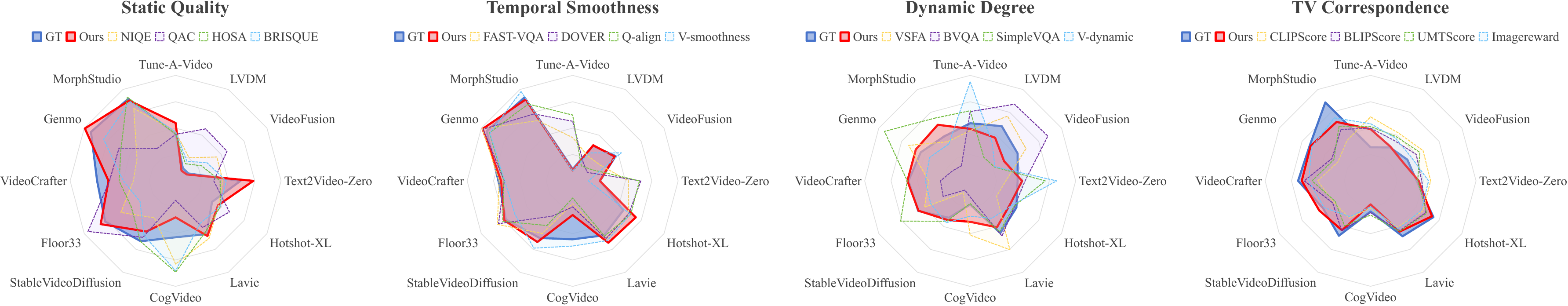}
 \vspace{-7mm}
	\caption{Comparison of win rates of different generation models across four dimensions evaluated by different VQA methods, demonstrating our AIGV-Assessor has better win-rate evaluation ability aligned with Ground Truth (GT). }
	\label{leida2}
\end{figure*}

%% file: table/ablation.tex
\begin{table*}
\vspace{-2mm}
\renewcommand\arraystretch{1.2}
  \caption{Ablation study of the proposed AIGV-Assessor method.}

  \vspace{-3mm}
  \resizebox{1\textwidth}{!}{
  \begin{tabular}{ccccc| ccc ccccccccc}
    \toprule
                                   & \multicolumn{4}{c}{Feature \& Strategy}                                  & \multicolumn{3}{c}{Static Quality}                   & \multicolumn{3}{c}{Temporal Smoothness}                  & \multicolumn{3}{c}{Dynamic Degree} &
                                   \multicolumn{3}{c}{T2V Correspondence} \\
    \cmidrule(r){2-5} \cmidrule(r){6-8} \cmidrule(r){9-11} \cmidrule(r){12-14}\cmidrule(r){15-17}
    \multicolumn{1}{c}{No.}        & spatial  & temporal   & quality level & LoRA finetuning    & SRCC     & PLCC     & KRCC          & SRCC       & PLCC    & KRCC          & SRCC     & PLCC   & KRCC          & SRCC      & PLCC   & KRCC      \\
    \multicolumn{1}{c}{(1)}          & \ding{52}   &           &      \ding{52}      &             & 0.864     & 0.866     & 0.726     & 0.870     & 0.868     & 0.727     & 0.556     & 0.572    & 0.432    & 0.616     & 0.620    & 0.492   \\
    \multicolumn{1}{c}{(2)}          &    \ding{52}          &  &           &       \ding{52}     & 0.874     & 0.876     & 0.723      & 0.875    & 0.876     & 0.736     & 0.558    & 0.573    & 0.431     & 0.723     & 0.734     & 0.533  \\
    \multicolumn{1}{c}{(3)}          & \ding{52}   & & \ding{52} & \ding{52}    & 0.887     & 0.884    & 0.722     & 0.881     & 0.883    & 0.706     & 0.562     & 0.575     & 0.433    & 0.739     & 0.758     & 0.544   \\
    \multicolumn{1}{c}{(4)}          &    \ding{52}         &  \ding{52}         & \ding{52} &              & 0.887     & 0.888    & 0.753     & 0.917    & 0.910    & 0.796    & 0.569     & 0.536     & 0.438     & 0.688     & 0.673     & 0.557  \\
    \multicolumn{1}{c}{(5)}          & \ding{52}   & \ding{52} &           &     \ding{52}       & 0.905     & 0.908     & 0.754     & 0.919     & 0.917     & 0.799     & 0.589     & 0.587     & 0.441     & 0.742     & 0.763    & 0.549   \\
    
   \rowcolor{gray!20} \multicolumn{1}{c}{(6)}          & \ding{52}   & \ding{52} & \ding{52} & \ding{52}  &\textbf{ 0.916 }    & \textbf{0.919}     & \textbf{0.758 }    &\textbf{ 0.923 }    & \textbf{0.922 }    & \textbf{0.804     }& \textbf{0.609     }& \textbf{0.608   }  & \textbf{0.444 }    & \textbf{0.750 }    & \textbf{0.770  }   & \textbf{0.559 }  \\
    \bottomrule
  \end{tabular}\label{ablation}
   }
  \vspace{-5mm}
  \centering
\end{table*}

%% file: section/6_conclusion.tex
 \vspace{-1mm}
\section{Conclusion}
 \vspace{-1mm}
In this paper, we study the human visual preference evaluation problem for AIGVs. We first construct AIGVQA-DB, which includes 36,576 videos generated based on 1048 various text-prompts, with the MOSs and pair comparisons evaluated from four perspectives.
Our detailed manual evaluations reflect different aspects of human visual preferences on AIGVs and reveal critical insights into the strengths and weaknesses of various text-to-video models.
Based on the database, we evaluate the performance of state-of-the-art quality evaluation models and establish a new benchmark, revealing their limitations in measuring the perceptual preference of AIGVs.
Finally, we propose AIGV-Assessor, a novel VQA model that leverages the capabilities of LMMs to give quality levels, predict quality scores, and compare preferences from four dimensions. 
Extensive experiments demonstrate that AIGV-Assessor achieves state-of-the-art performance on both AIGVQA-DB and other AIGVQA benchmarks, validating its robustness in understanding and evaluating the AI-generated videos.

%% file: supple/0_importance.tex
\section{Significance of AIGVQA-DB Construction}
\input{figure/sig}

Mean opinion scores (MOS) have traditionally served as the primary metric for measuring overall quality. While MOS is effective for providing a general indication of quality, it has notable limitations, especially when it comes to high-quality content. For example, when evaluating closely matched, high-quality images or videos, MOS often results in similar scores across samples, leading to coarse evaluations that fail to capture subtle differences in factors like exposure, motion smoothness, or color fidelity.
As illustrated in Figure \ref{sig}, human assessors applying absolute scoring frequently yield inconsistent ratings due to varying personal standards or subjective preferences. Despite this, when asked to make relative comparisons—such as deciding whether "video1 is better than video2"—they exhibit greater consistency and are able to reach a reliable consensus. This highlights a crucial insight: relative comparisons offer more precision and consistency than absolute scoring alone. Pairwise comparisons, which focus on directly comparing two samples, have thus emerged as a valuable complement to MOS. By emphasizing relative differences, pairwise assessments allow for finer granularity, capturing nuanced distinctions that absolute scores may miss. Numerous studies have demonstrated the effectiveness of pairwise comparisons in reducing ambiguity in scoring and providing more detailed evaluations, especially when assessing high-quality content \cite{prashnani2018pieapp, zhang2018unreasonable, zhang2019ranksrgan, wu2023better, kirstain2023pick}.

The development of AI-generated image quality assessment (AIGIQA) datasets is already relatively well-established, incorporating both MOS for absolute quality evaluation and pairwise comparisons for assessing relative quality differences. This dual approach has proven effective in capturing both the overall and relative quality aspects of AI-generated images. However, existing AI-generated video quality assessment (AIGVQA) datasets primarily rely on MOS alone, which significantly limits their ability to capture the fine-grained quality differences inherent in video content. Videos, unlike static images, present unique challenges such as temporal coherence, spatial consistency, and the dynamic nature of objects and motion. These challenges make pairwise comparisons particularly valuable in video quality assessment, as they allow for more accurate evaluations of complex attributes like motion smoothness and frame-to-frame consistency. Unfortunately, most current VQA datasets lack systematic, large-scale pairwise data, limiting their capacity to assess these dynamic and temporal aspects effectively.
To address this gap, we propose the AIGVQA-DB dataset, including both MOS and pairwise comparison data, offering a robust framework for evaluating the intricate nuances of AI-generated video content. By incorporating both absolute quality judgments and relative assessments, AIGVQA-DB enables more accurate model training and evaluation, allowing models to learn not only the absolute quality standards but also the relational nuances inherent in video sequences. 

%% file: figure/sig.tex
\begin{figure}[!t]
	\centering
	\includegraphics[width=\linewidth]{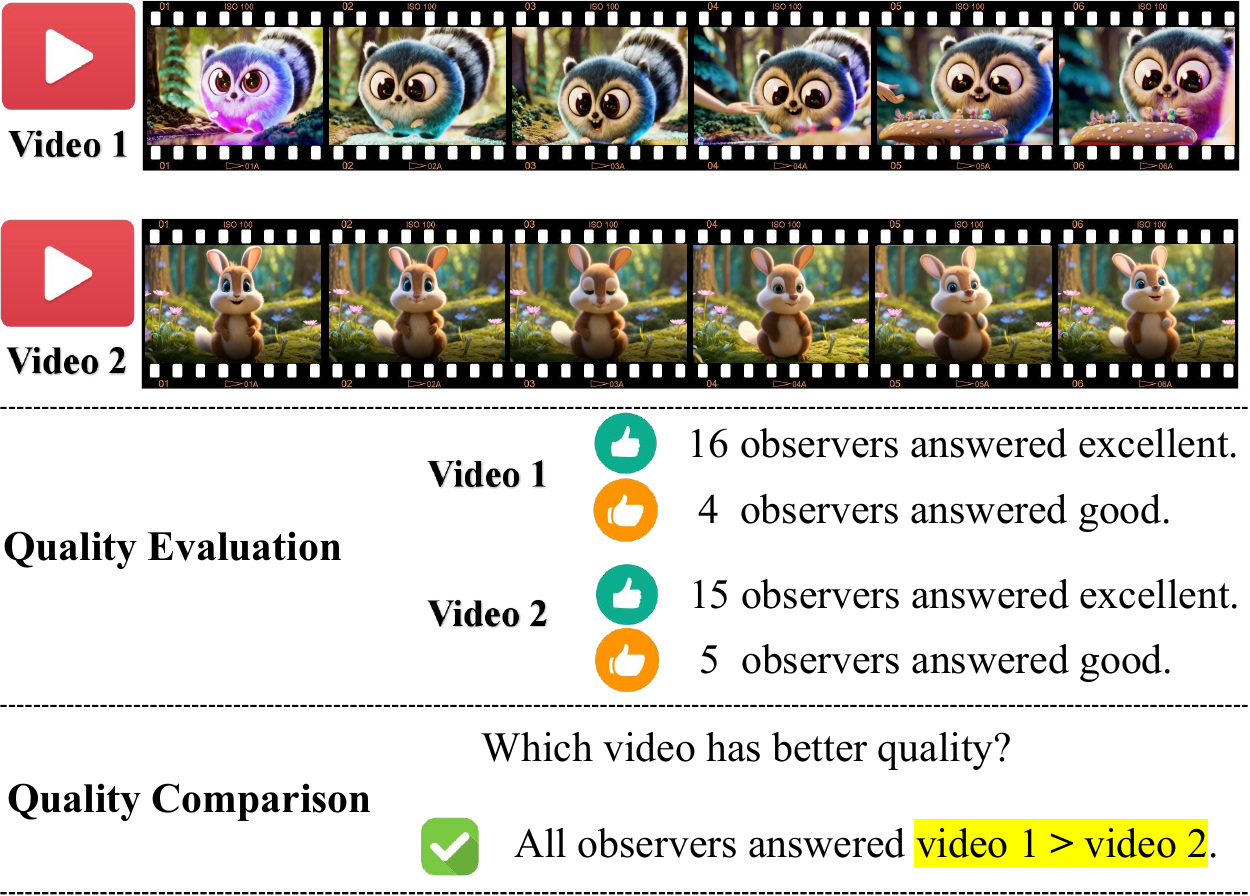}
 \vspace{-6mm}
	\caption{The motivation for visual quality comparison: single stimulus absolute ratings like "excellent" or "good" often involve randomness or inconsistency due to varying personal standards. Double stimuli comparative settings avoid the ambiguity of absolute evaluations for single videos, providing clearer and more consistent judgments.}

 \vspace{-3mm}
	\label{sig}
\end{figure}

%% file: supple/1_datadetails.tex
\input{table/prompt_description}
\section{More Details of Video Generation}
\label{B}
\subsection{Detailed Information of Prompts}
The AIGVQA-DB dataset offers a rich and diverse collection of prompts, carefully constructed from two sources: (1) existing open-domain text-video pair datasets, including InternVid \cite{wang2023internvid}, MSRVTT \cite{xu2016msr-vtt}, WebVid \cite{Bain21}, TGIF \cite{li2016tgif}, FETV \cite{liu2024fetv} and Sora website \cite{videoworldsimulators2024}. These datasets contribute a robust foundation of real-world and generalizable scenarios, providing a solid basis for training and evaluation. (2) manually written prompts designed to push the boundaries of model robustness and generalization. Inspired by unique categories such as "Imagination" and "Conflicting," these prompts introduce rare or non-realistic scenarios, like "A panda is flying in the sky," that test a model’s ability to handle creative and unconventional inputs. 
As illustrated in Table \ref{pc}, each prompt in our dataset is categorized based on four key aspects, including “spatial major content”,  “temporal major content”, “attribute control”, and “prompt complexity”. For each aspect, we include typical elements that frequently occur in daily life. As shown in Figure 5, the spatial major content focuses on objects described in the prompt, including ten subcategories: people, animals, plants, and \textit{etc.} In contrast, temporal major content highlights dynamic actions or changes and is divided into four subcategories, including actions, kinetic motions, fluid motions, and light change, as shown in  Figure 6. Similarly, the attribute control covers specific stylistic or compositional controls embedded in prompts, enabling nuanced customization of generated content, including color, quantity, camera view, speed, motion direction, and event order, as shown in  Figure 7. Additionally, we classify prompt complexity into three levels including: simple, medium, and complex, based on the number of descriptive elements in the text. By integrating real-world scenarios, imaginative constructs, and a structured categorization system, AIGVQA-DB ensures a comprehensive evaluation framework that challenges text-to-video generation models in both realistic and highly creative contexts.

To ensure a comprehensive and systematic classification of prompts within the AIGVQA-DB dataset, we employed the GPT-4 API for multi-aspect prompt categorization. 
GPT-4 was provided with task-specific instructions designed to guide its classification process. These instructions included detailed descriptions of the categorization task, along with illustrative examples to ensure consistent and accurate labeling. Prompts were analyzed based on key aspects. For instance, to classify spatial major content, GPT-4 was prompted with a detailed instruction template, such as:
\begin{tcolorbox}[colback=white, colframe=black] \textit{"Analyze the following prompt and classify it into one or more categories based on the type of object it describes. The categories include People, Buildings and Infrastructure, Animals, Artifacts, Vehicles, Plants, Scenery and Natural Objects, Food and Beverage, and Illustration. Provide only the category names as the output. Example: 'A cat is sitting under a tree.' Spatial major content: Animals, Plants."} \end{tcolorbox}
\input{table/generation}
Under this framework, GPT-4 processes the given prompt and assigns appropriate category labels based on its analysis. For example, a prompt like "A dog is driving a car." would be classified under Animals and Vehicles. Similar categorization instructions were devised for temporal major content and attribute control.
Additionally, prompt complexity is classified based on the number of non-stop words present in each prompt. This multi-aspect categorization approach ensured that every prompt in the AIGVQA-DB was exhaustively labeled, facilitating fine-grained evaluation of text-to-video models. Examples of prompts and their corresponding categorizations in AIGVQA-DB are shown in Table \ref{prompts}.


\subsection{Detailed Information of Text-to-Video Models}
To construct AIGVQA-DB, we utilize 15 state-of-the-art text-to-video generative models, encompassing both open-source and closed-source methods, as detailed in Table \ref{models}. For open-source models, we rely on official repositories and use default weights to standardize results and maintain consistency across experiments. For closed-source models, we leverage publicly available APIs from open-source platforms.  This comprehensive selection ensures that AIGVQA-DB serves as a robust benchmark for evaluating text-to-video generation systems.

\noindent
{\bf CogVideo.}
CogVideo~\cite{hong2022cogvideo} is built on the text-to-image model CogView2~\cite{ding2022cogview2}. It employs a multi-frame-rate hierarchical training strategy to ensure better alignment between text and temporal counterparts in videos, generating keyframes based on textual prompts and recursively interpolating intermediate frames for coherence. 

\noindent
{\bf LVDM.} 
LVDM~\cite{he2022latent} is an efficient video diffusion model operating in a compressed latent space, designed to address the computational challenges of video synthesis. It uses a hierarchical framework to extend video generation beyond training lengths, effectively mitigating performance degradation via conditional latent perturbation and unconditional guidance techniques. 

\noindent
{\bf Tune-A-Video.} 
Tune-A-Video~\cite{wu2023tune} is a one-shot text-to-video generation model that extends text-to-image (T2I) models to the spatio-temporal domain. It uses sparse spatio-temporal attention to maintain consistent objects across frames, overcoming computational limitations. It can synthesize novel videos from a single example compatible with personalized and conditional pretrained T2I models. 

\noindent
{\bf VideoFusion.} 
VideoFusion \cite{luo2023videofusion} is a decomposed diffusion probabilistic model for video generation. Unlike traditional methods that add independent noise to each frame, it separates noise into shared base noise and residual noise, improving spatial-temporal coherence. This approach leverages pretrained image-generation models for efficient frame content prediction while maintaining motion dynamics. 

\noindent
{\bf Text2Video-Zero.} Text2Video-Zero \cite{khachatryan2023text2video} is a zero-shot text-to-video synthesis model without any further fine-tuning or optimization, which introduces motion dynamics between the latent codes and cross-frame attention mechanism to keep the global scene time consistent. We adopt its official code with default parameters (\texttt{<motion\_field\_strength\_x\&y=12>}).

\noindent
{\bf LaVie.} LaVie \cite{wang2023lavie} is an integrated video generation framework that operates on cascaded video latent diffusion models. For each prompt, we use the base T2V model and sample 16 frames of size 512$\times$320 at 8 FPS. The number of DDPM \cite{ho2020denoising} sampling steps and guidance scale are set as 50 and 7.5, respectively.

\noindent
{\bf VideoCrafter.} VideoCrafter \cite{chen2023videocrafter1} is a video generation and editing toolbox. We sample 16 frames of size 1024$\times$576 at 8 FPS, according to its default settings. 

\noindent
{\bf Hotshot-XL.} Hotshot-XL \cite{Mullan_Hotshot-XL_2023} is a text-to-gif model trained to work alongside Stable Diffusion XL\footnote{\url{https://huggingface.co/hotshotco/SDXL-512}}. We adopt its official code with default parameters and change the output format from \texttt{GIF} to \texttt{MP4}.

\noindent
{\bf Genmo.} Genmo \cite{gem} is a high-quality video generation platform. We generate 60 frames of size $\leq$2048$\times$1536 at 15 FPS for each prompt. The motion parameter is set to 70\%.

\noindent
{\bf Gen-2.} Gen-2 \cite{gen} is a multimodal AI system, introduced by Runway AI, Inc., which can generate novel videos with text, images or video clips. We collect 96 frames of size 1408$\times$768 at 24 FPS for each prompt. 

\noindent
{\bf Sora.} Sora~\cite{videoworldsimulators2024} is particularly known for its ability to handle complex, multi-element prompts, ensuring coherent visual representations of diverse scenarios. Sora~\cite{videoworldsimulators2024} currently does not have an open-source API, so the videos we used are downloaded from its official website.

\noindent
{\bf Floor33, MoonValley and MorphStudio.} Floor33 \cite{floor33}, MoonValley \cite{moonvalley}, and MorphStudio \cite{morph} are recent popular online video generation application. We use the T2V mode of these applications via commands in Discord\footnote{\url{https://discord.com}}. 

%% file: table/prompt_description.tex
\begin{table*}
\caption{Prompt categorizations with subcategories, detailed descriptions, and representative keyword examples. }
\label{pc}
\resizebox{\textwidth}{!}{
    \begin{tabular}{clllllll}
    \toprule
   Category   & Subcategory  & Descriptions & Keyword examples\\
    
    \midrule               
    \multirow{9}{*}{Spatial major content}   
    & People &  Prompts that include humans. &  person, man, woman, men, women, kid, girl, boy, baby   \\
    & Plants &  Prompts that include plants. & flower, leaf, tree, grass, forest, wheat, plant, peony          \\
    & Animals &  Prompts that include animals. & panda, dog, cat, elephant, horse, bird, butterfly, rabbit     \\
    & Vehicles &  Prompts that include vehicles. & car, van, plane, tank, carriage, rocket, motorcycle      \\
    & Artifacts & Prompts that include human-made objects. & robot, doll, toy, microphone, paper, plate, bowl, ball       \\
    & Illustrations & Prompts that include geometrical objects and symbols. & abstract, pattern, particle, gradient, loop, graphic, line    \\
    & Food and beverage  & Prompts that include food and beverage. & water, wine, coffee, apple, butter, egg, chocolate, lime     \\
    & Buildings and infrastructure  & Prompts that include buildings and infrastructure.   & room, building, bridge, court, concert, hotel, factory   \\
    & Scenery and natural objects   & Prompts that include lifeless natural objects and scenery. & wind, sand, snow, rain, sky, fog, mountain, river, sun  \\
    \midrule                
    \multirow{4}{*}{Temporal major content}  
    & Actions & Prompts that include the motion of solid objects & sing, dance, laugh, cry, smile, jump, walk, eat, drink \\
    & Kinetic motions& Prompts that include the motion of solid objects. & fly, spin, race, move, rotate, fall, rise, bounce, sway\\
    & Fluid motions &Prompts that include the motions of fluids or like fluids.& waterfall, wave, fountain, smoke, steam, inflate, melt\\
    & Light change &Prompts from which the generated videos may involve light change. &sunset, sunrise, firework, shine, glow, burn, flash, bright \\
    \midrule                
    \multirow{6}{*}{Attribute control} 
    & Color & Prompts that include colors. & white, pink, black, red, green, purple, blue, yellow\\
    & Quantity & Prompts that include numbers. & one, two, three, four, five, six, seven, eight, nine, ten\\
    & Camera view & Prompts that include control over the camera view. &view, macro, film, close, capture, aerial, shot, camera\\
    & Speed & Prompts that include control over speed. & fast, slow, rapid, speed, motion, time, quick, swift, lag\\
    & Event order & Prompts that include control over the order of events. & then, before, after, first, second \\
    & Motion direction & Prompts that include control over the motion direction. & forward, backward, from, into, through, out of, left, right\\

    \midrule
     \multirow{3}{*}{Prompt complexity} & Simple & Prompts that involve 0 $\sim$ 8 non-stop words. & -\\

      & Medium & Prompts that involve 9 $\sim$ 11 non-stop words. & -\\
      & Complex & Prompts that involve more than 11 non-stop words. & -\\
    \bottomrule
    \end{tabular}
}
\centering
\end{table*}

%% file: table/generation.tex
\begin{table*}
\centering
  \caption{Video formats and numbers generated by the 15 text-to-video (T2V) models in the AIGVQA-DB. $\checkmark $ in the Pairs and MOS columns indicate which generative models are utilized in each of the two subsets.  \textsuperscript{$\dag $} Representative variable. *Representative open-source.}
  \label{models}
  \vspace{-2mm}
    \renewcommand\arraystretch{1.1}
  \resizebox{\textwidth}{!}{\begin{tabular}{lcccccccl}
    \toprule
     Models                         & Number & Prompts &Frames &FPS  & Resolution & MOS &Pairs & URL\\
    \midrule 
     *CogVideo~\cite{hong2022cogvideo} & 4,000 & 1,000 & 32 & 10  & 480$\times$480 & - & $\checkmark $  &\url{https://github.com/THUDM/CogVideo}\\
     *LVDM~\cite{he2022latent}         & 4,048 & 1,048 & 16 & 8   & 256$\times$256 & $\checkmark $ & $\checkmark $ &\url{https://github.com/YingqingHe/LVDM}\\
     *Tune-A-Video~\cite{wu2023tune}   & 4,048 & 1,048& 8  & 8   & 512$\times$512 & $\checkmark $ & $\checkmark $ &\url{https://github.com/showlab/Tune-A-Video}\\
     *VideoFusion~\cite{luo2023videofusion} & 4,048 & 1,048& 16  & 8  & 256$\times$256 & $\checkmark $ & $\checkmark $ &\url{https://github.com/modelscope/modelscope}\\
     *Text2Video-Zero \cite{khachatryan2023text2video}&4,048 & 1,048&  8 & 4 & 512$\times$512 & $\checkmark $ & $\checkmark $ &\url{https://github.com/Picsart-AI-Research/Text2Video-Zero}\\
     *Lavie \cite{wang2023lavie}&4,048 & 1,048& 16 & 8 & 512$\times$320 & $\checkmark $ & $\checkmark $ &\url{https://github.com/Vchitect/LaVie}\\
     *VideoCrafter \cite{chen2023videocrafter1} & 4,048 & 1,048& 16 & 10 & 1024$\times$576& $\checkmark $ & $\checkmark $ &\url{https://github.com/AILab-CVC/VideoCrafter}\\
     *Hotshot-XL \cite{Mullan_Hotshot-XL_2023} & 4,048 & 1,048 & 8 & 8 &672$\times$384 & $\checkmark $ & $\checkmark $&\url{https://github.com/hotshotco/Hotshot-XL}\\
     *StableVideoDiffusion \cite{blattmann2023stable} & 1,000 & 1,000& 14 & 6 &576$\times$1024 & - &$\checkmark $ &\url{https://github.com/Stability-AI/generative-models}\\
     \hdashline
     Floor33 \cite{floor33} & 4,048 & 1,048& 16 & 8 &1024$\times$640 & $\checkmark $ & $\checkmark $&\url{https://discord.com/invite/EuB9KT6H}\\
     Genmo \cite{gem}&4,048 & 1,048& 60 & 15 & 2048$\times$1536\textsuperscript{$\dag $} & $\checkmark $ & $\checkmark $&\url{https://www.genmo.ai}\\
     Gen-2 \cite{gen}&48 & 48& 96 & 24& 1408$\times$768 & $\checkmark $ & - &\url{https://research.runwayml.com/gen2}\\
     MoonValley \cite{moonvalley} & 48 & 48& 200\textsuperscript{$\dag $} & 50 & 1184$\times$672 & $\checkmark $ & - &\url{https://moonvalley.ai}\\
     MorphStudio \cite{morph}&4,000 & 1,000& 72 & 24 &1920$\times$1080 & - & $\checkmark $&\url{https://www.morphstudio.com}\\  
     Sora~\cite{videoworldsimulators2024} & 48 & 48 &600\textsuperscript{$\dag $} & 30  &1920$\times$1080\textsuperscript{$\dag $} & $\checkmark $ & - &\url{https://openai.com/research}\\
    \bottomrule
  \end{tabular}}
    \vspace{-3mm}
\end{table*}

%% file: supple/2_subjective.tex
\section{More Details of Subjective Experiment}

\subsection{Annotaion Criteria}
The assessment criteria for AIGVQA-DB are systematically structured across four key dimensions: static quality, temporal smoothness, dynamic degree, and text-video correspondence. These dimensions provide a comprehensive framework for video quality assessment, ensuring thorough and reliable assessments through clearly defined scales, detailed annotation criteria, and illustrative reference examples.
\begin{itemize}
\item \textbf{Static quality} focuses on the video's visual clarity, naturalness, color balance, and detail richness. High-scoring videos are characterized by exceptional clarity, vivid and well-balanced colors, and meticulous attention to detail, offering an immersive and visually striking experience. Conversely, low scores reflect videos with blurriness, unnatural color tones, faded visuals, and lack of clarity or detail. This dimension captures the foundational visual attributes that make a video aesthetically pleasing or distracting. For detailed criteria, refer to Figure \ref{sup_bz1}.
\item \textbf{Temporal smoothness} evaluates the consistency and fluidity of frame-to-frame transitions, and the naturalness of object movements within the video. Videos with high scores exhibit seamless transitions, smooth movements, and no noticeable inconsistencies, creating a natural and immersive viewing experience. Low scores denote irregular or abrupt frame changes and disjointed object movements, which detract from the overall fluidity. For detailed criteria, refer to Figure \ref{sup_bz2}.
\item \textbf{Dynamic degree} assesses the range and expressiveness of motion within the video. High-scoring videos display diverse, realistic, and natural movements of objects, animals, or humans, contributing to a vivid and engaging experience. Lower scores indicate limited motion or unnatural dynamics. This dimension highlights the importance of motion diversity and realism in engaging content. For detailed criteria, refer to Figure \ref{sup_bz3}.
\item \textbf{Text-video correspondence} examines the alignment between the video content and its associated text prompt. Videos with high scores perfectly match the descriptions in the prompt, accurately reflecting all elements with high fidelity. These videos effectively translate textual information into visual content without omissions or mismatches. In contrast, videos with lower scores exhibit inconsistencies, missing elements, or mismatched content. For detailed criteria, refer to Figure \ref{sup_bz4}.
\end{itemize}
Each of these four dimensions is supported by detailed examples, providing annotators with clear guidelines to perform evaluations. This systematic approach ensures accuracy and consistency in the annotation process, enabling a robust analysis of human preference and video quality. 
\subsection{Annotation Interface}
To ensure a comprehensive and efficient evaluation of video quality, we designed two custom annotation interfaces tailored for different assessment tasks: one for score annotation and the other for pair annotation. The score annotaion interface, shown in Figure \ref{UIscore}, is a manual evaluation platform developed using the Python tkinter package, designed to facilitate MOS assessments. To ensure uniformity and minimize resolution-related biases in video quality evaluation, all videos displayed in this interface are cropped to a spatial resolution of 512×512 pixels. The duration of the videos remains unaltered, preserving the full content described in the associated text prompts. Meanwhile, the pair annotation interface, illustrated in Figure \ref{UIpair}, supports paired comparison assessments, where participants evaluate two videos side-by-side. This interface is designed to explore preference judgments across four key aspects, including: static Quality, temporal Smoothness, dynamic Degree, and text-Video Correspondence. In each comparison, participants are shown two videos, labeled "A" and "B," and are required to select their preferred video for each aspect. The interface ensures an unbiased evaluation environment by clearly distinguishing between the two videos while allowing side-by-side playback. Participants can replay either video as needed before making their selection. The evaluation process emphasizes subjective preferences while offering a structured approach to gather comparative insights across multiple dimensions. Navigation options, such as "Replay," "Next," and "Save," streamline the workflow, enabling efficient annotation


\input{figure/UIscore}
\input{figure/UIpair}
\subsection{Annotation Management}
To ensure ethical compliance and the quality of annotations, we implemented a comprehensive process for the AIGVQA-DB dataset. All participants were fully informed about the experiment's purpose, tasks, and ethical considerations. Each participant signed an informed consent agreement, granting permission for their subjective ratings to be used exclusively for non-commercial research purposes. The dataset, consisting of 36,576 AI-generated videos (AIGVs) and their associated prompts, is publicly released under the CC BY 4.0 license.
We ensured the exclusion of all inappropriate or NSFW content (textual or visual) through a rigorous manual review during the video generation stage. 
The annotation was divided into two key components: paired comparison annotation and MOS annotation, each designed to evaluate videos across four dimensions, including: static quality, temporal smoothness, dynamic degree, and text-video correspondence.
For the paired comparisons, 30,000 video pairs were evaluated by a total of 100 participants. Each pair was assessed by three participants, and the final result for each pair was determined by majority voting. In cases of discrepancies, the average opinions of the three participants were calculated to resolve the tie. This approach ensured a balanced and fair evaluation of preferences between video pairs.
The MOS annotation task involved 20 participants to rate all videos in the MOS subset individually. Participants scored each video on a 0-5 Likert scale across the four evaluation dimensions. This granular scoring provided a comprehensive dataset for analyzing human preferences and video quality.

Before participating in the annotation tasks, all participants underwent a rigorous training process. They were provided with detailed instructions, multiple standard examples (Figures \ref{sup_bz1}-\ref{sup_bz4}), and step-by-step guidance on the annotation criteria. A pre-test was conducted to evaluate participants’ understanding of the criteria and their agreement with standard examples. Those who did not meet the required accuracy were excluded from further participation.
During the experiment, all evaluations were conducted in a controlled laboratory environment with normal indoor lighting. Participants were seated at a comfortable viewing distance of approximately 60 cm from the screen. To further reduce potential biases, videos from different models were alternately presented in the both MOS and pair comparison tasks.
Although individual preferences may vary, the use of detailed explanations and standardized annotation criteria ensured a high level of agreement across participants. This consensus was particularly evident in pair annotations, where majority voting captured group preferences effectively. The documentation of the entire annotation process served as a reference and training standard, ensuring consistency and reliability across all evaluations.
This rigorous annotation management strategy makes AIGVQA-DB a robust and ethically sound resource for advancing research in video quality assessment.

\section{More Details of AIGVQA-DB}
\label{D}
\subsection{Detailed Information of the Subsets}
\paragraph{Construction of the MOS subset.}
The MOS subset is specifically designed to evaluate the perceptual quality of videos generated by T2V models, offering a comprehensive benchmark for subjective evaluation. This subset incorporates contributions from 12 generative models in the database, encompassing a broad spectrum of temporal and spatial attributes to ensure diversity.
To construct this subset, we initially sourced 48 high-quality videos and their corresponding textual prompts from the Sora platform~\cite{videoworldsimulators2024}. These prompts were then used to generate additional videos utilizing 11 other generative models, resulting in a total of 576 videos (48 prompts × 12 generative models). This approach ensured the inclusion of a wide range of visual styles and generative qualities. The dataset spans significant variations in frame count, frame rate (FPS), and resolution, ranging from the compact 256×256 outputs of VideoFusion~\cite{luo2023videofusion} at 8 FPS to the high-definition 1920×1080 outputs of Sora at 30 FPS. Such diversity in video attributes allows for a robust analysis of generative models under different visual and temporal conditions.
Each video in the MOS subset is evaluated by 20 annotators across four dimensions: Static Quality, Temporal Smoothness, Dynamic Degree, and Text-Video Correspondence. This rigorous evaluation process results in 46,080 individual ratings (4 dimensions × 576 videos × 20 annotators). The annotators, equipped with detailed training and examples, provide subjective scores on a 0-5 Likert scale, ensuring consistency and reliability in their assessments.
By including videos with diverse visual properties, the MOS subset provides a robust foundation for subjective evaluation tasks, enabling researchers to compare T2V models based on the perceptual quality of AIGVs.

\input{figure/leida2}
\paragraph{Construction of the Pair comparison subset.}
To enable detailed comparative analysis, we construct the pair comparison subset.
This subset is built based on 1,000 carefully curated textual prompts, including a wide range of scenarios, themes, and levels of complexity. These prompts ensure diversity in content and provide a robust basis for assessing the performance of generative models across various contexts.
We use 12 generative models, including 8 open-source models such as Hotshot-XL~\cite{Mullan_Hotshot-XL_2023} and Floor33~\cite{floor33}, and 4 closed-source models, such as Gen-2~\cite{gen} and MoonValley~\cite{moonvalley}. For each prompt, open-source models generate four distinct videos, capturing variations in their generative outputs and showcasing intra-model diversity. Closed-source models, due to access constraints, produce one video per prompt. This comprehensive approach results in a dataset of 36,000 videos (1,000 prompts × (8 open-source models × 4 videos + 4 closed-source models × 1 video)). The videos in this subset exhibit a wide range of resolutions and frame counts, from the lower-resolution 256×256 outputs of LVDM~\cite{he2022latent} to the high-definition 1920×1080 videos from MorphStudio~\cite{morph}. 
Each video pair is evaluated by three annotators, who provide individual ratings for all four dimensions. These ratings are aggregated to determine the final result for each dimension, with majority voting or averaged scores used to resolve any disagreements. This process results in 360,000 (4 × 30,000 × 3) ratings, ensuring a rigorous and nuanced analysis of model performance.
The pair-comparison subset allows for head-to-head comparisons of generative models, and provides valuable insights into the strengths and weaknesses of different models, offering researchers a robust foundation for comparative studies.


\subsection{More Result Analysis}
We analyze the subjective pair ratings by calculating the win rates of different generation models across different categories, revealing strengths and weaknesses from four different dimensions.
For the evaluation of spatial content categories, as shown in Figure \ref{leida33}(a), models like Genmo \cite{gem} perform exceptionally well in generating realistic representations of people, animals, and vehicles showcasing their strong attention to detail and visual fidelity. MorphStudio \cite{morph} consistently leads in producing high-quality outputs for scenery and natural objects, excelling in generating visually appealing and immersive natural environments. Additionally, StableVideoDiffusion \cite{blattmann2023stable} demonstrates notable strength in creating illustrations, highlighting its flexibility in handling stylized and artistic content. Conversely, LVDM ~\cite{he2022latent} and VideoFusion \cite{luo2023videofusion} lag in these categories, struggling with resolution and detail preservation.
For the evaluation of temporal content categories, as shown in Figure \ref{leida33}(b), MorphStudio \cite{morph} excels in handling kinetic motions, fluid motions, actions, and scenarios with light changes, making its outputs maintain high temporal smoothness and text-video correspondence. However, models like Text2Video-Zero \cite{khachatryan2023text2video} occasionally produce abrupt transitions, and Tune-A-Video~\cite{wu2023tune} shows limitations in maintaining temporal smoothness under complex motion conditions.
For the evaluation of attribute control categories, as shown in Figure \ref{leida33}(c),
Genmo \cite{gem} performs well in maintaining appropriate quantities of objects. Floor33 \cite{floor33} and VideoCrafter \cite{chen2023videocrafter1} display superior performance in the logical sequence of events. In contrast, StableVideoDiffusion~\cite{blattmann2023stable} encounters challenges in event order. Its generative process involves first creating static images and subsequently animating them to produce video sequences. The static-to-dynamic generation pipeline introduces discrepancies in temporal alignment, making it difficult to ensure that actions unfold in a logically consistent manner.
For the evaluation of prompt complexity categories, as shown in Figure \ref{leida33}(d),
most models demonstrate competence in handling prompts of different complexity, likely due to shared architectures like diffusion-based systems, with common strengths and limitations in handling complex prompts.


%% file: figure/UIscore.tex
\begin{figure*}[!t]
	\centering
	\includegraphics[width=\linewidth]{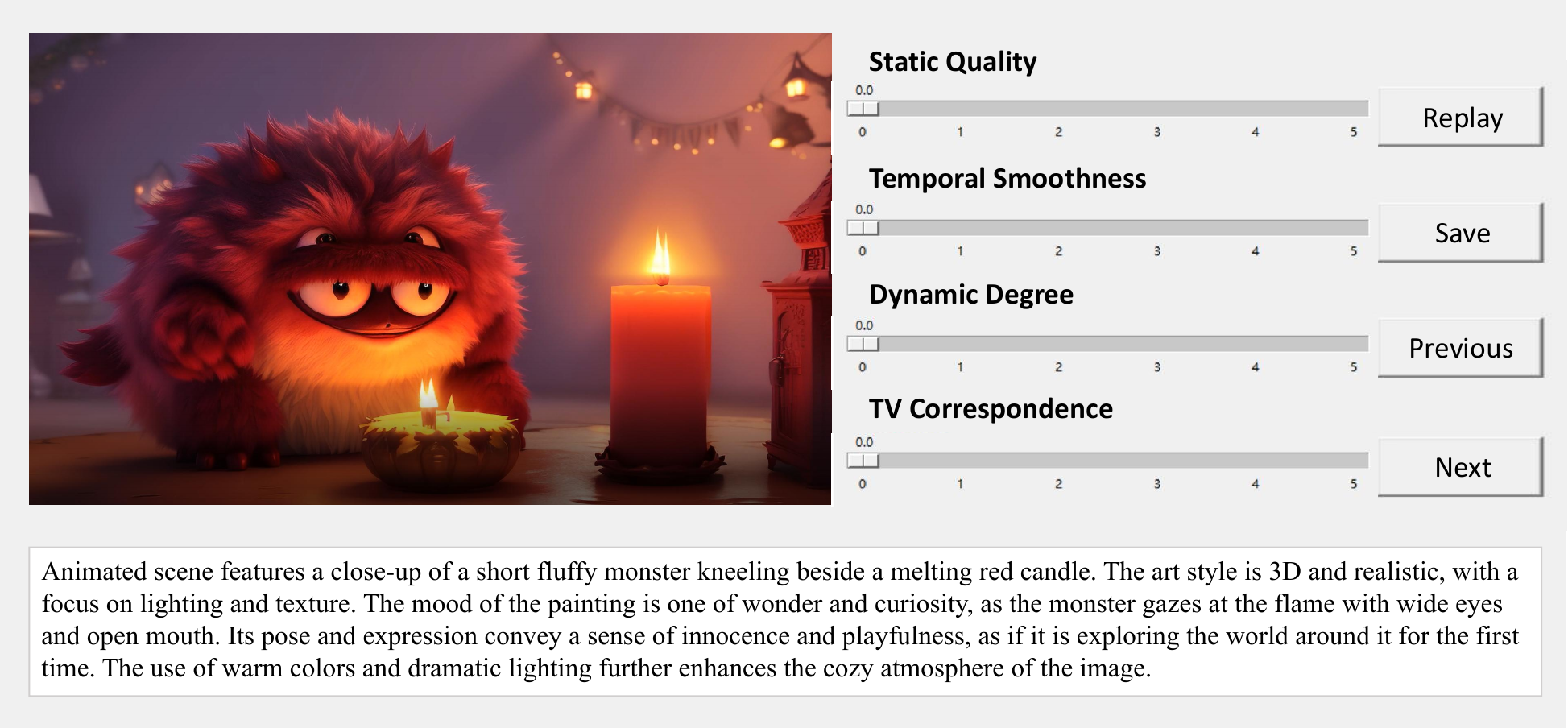}
	\caption{ An example of the rating assessment interface for human evaluation. The subjects are instructed to rate four dimensions of AI-generated videos, i.e., static quality, temporal smoothness, dynamic degree, and text-video correspondence, based on the given video and its prompt.}
	\label{UIscore}
\end{figure*}

%% file: figure/UIpair.tex
\begin{figure*}[!t]
	\centering
	\includegraphics[width=\linewidth]{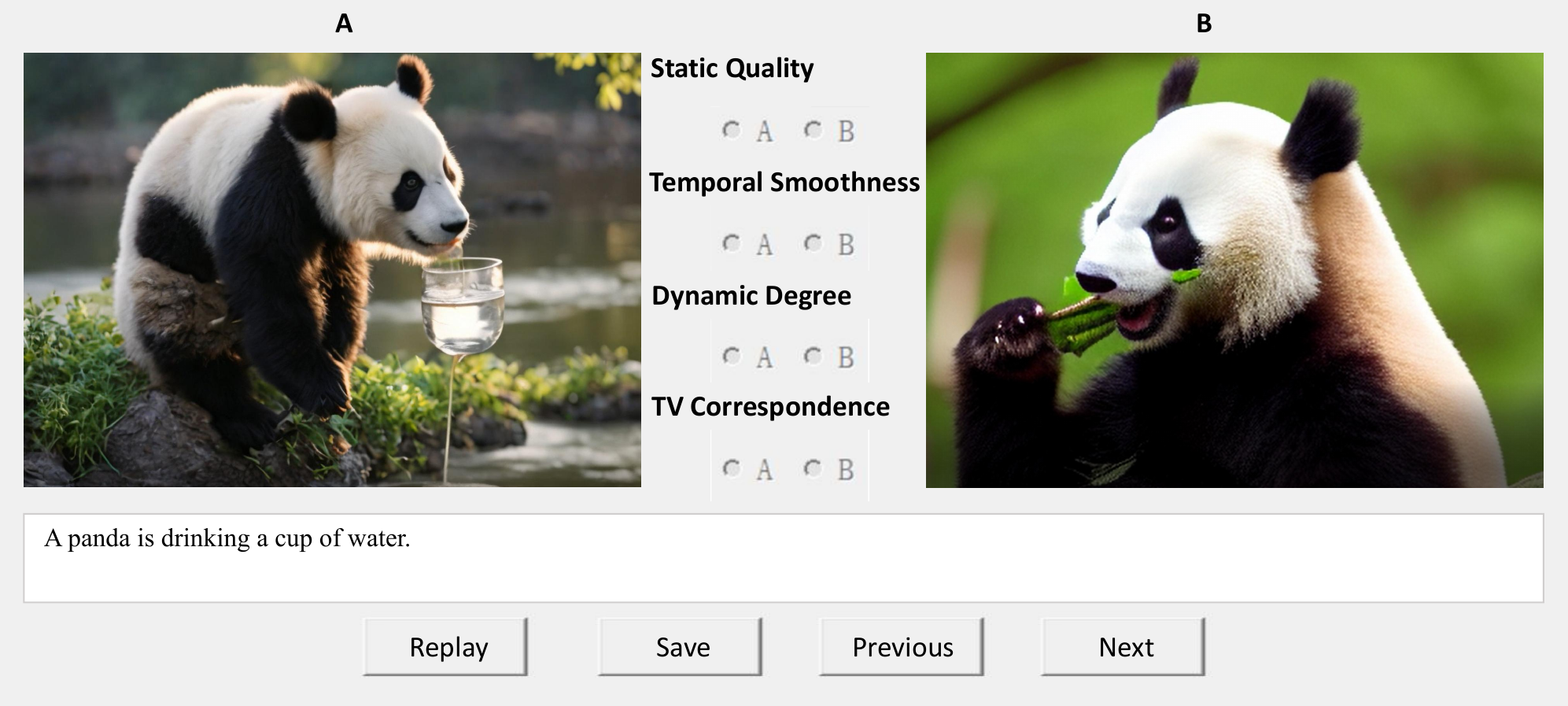}
	\caption{ An example of the pair comparison assessment interface for human evaluation. The subjects are instructed to choose which AI-generated video is better among the video pairs, considering four dimensions respectively, i.e., static quality, temporal smoothness, dynamic degree, and text-video correspondence.}
	\label{UIpair}
\end{figure*}

%% file: figure/leida2.tex
\begin{figure*}[!t]
	\centering
	\includegraphics[width=\linewidth]{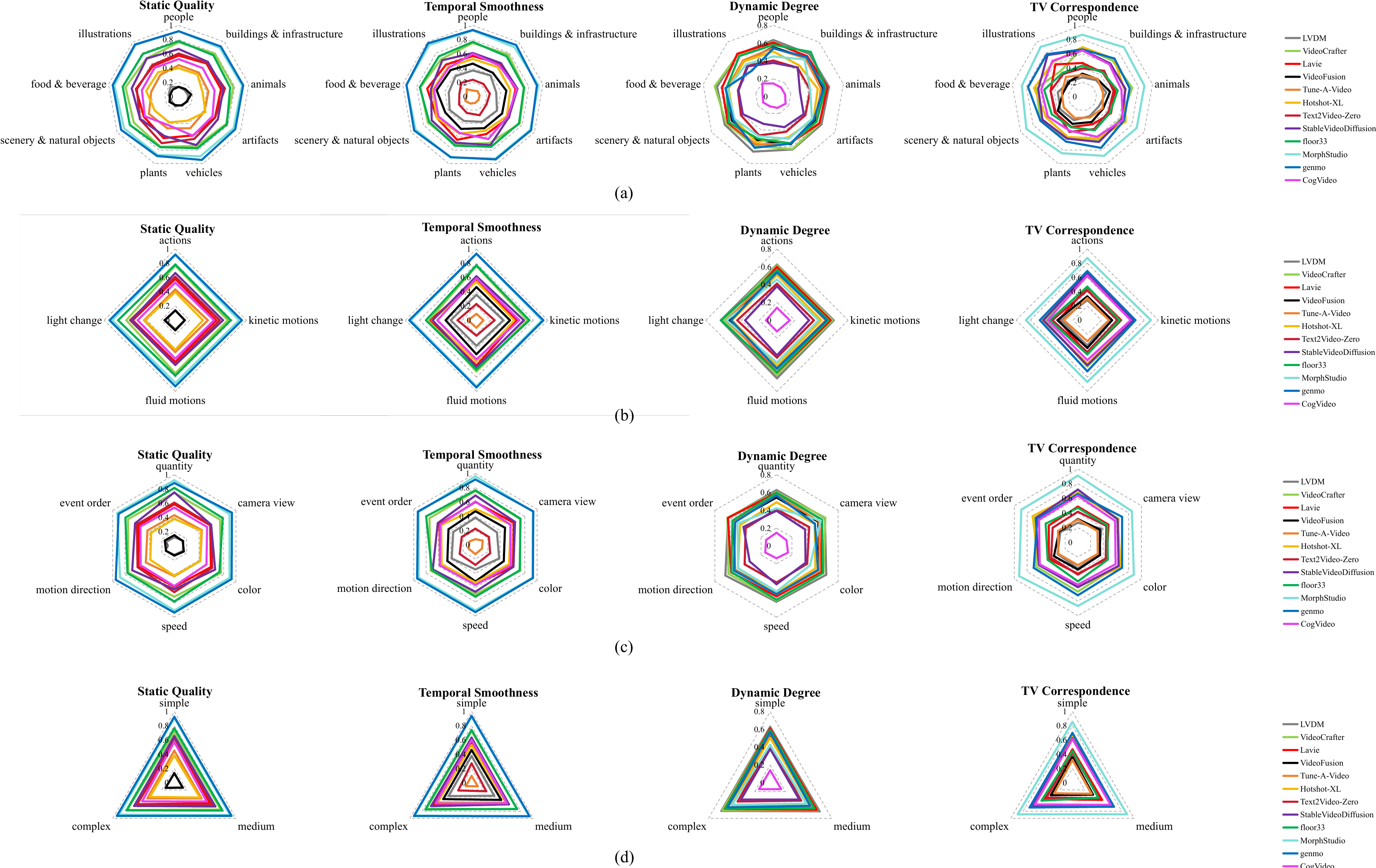}
 \vspace{-5mm}
	\caption{Comparison of averaged win rates of different generation models across different categories. (a) Results across spatial major contents. (b) Results across temporal major contents. (c) Results across dynamic degrees. (d) Results across text-to-video correspondence. }
 \vspace{-3mm}
	\label{leida33}
\end{figure*}

%% file: supple/3_Method.tex
\section{Details of Loss Function}
The training process for AIGV-Assessor is divided into three progressive stages, each utilizing a specific loss function to target distinct objectives: language loss for aligning visual and language features, L1 loss for generating accurate quality scores, and cross-entropy loss for robust pairwise video quality comparisons.

\paragraph{(1) Aligning visual and language features with language loss.}
In the first stage, spatial and temporal projectors are trained to align visual and language features using the language loss. This involves ensuring that the visual tokens extracted from the vision encoder correspond effectively to the language representations from the LLM. The language loss, calculated using a cross-entropy function, measures the model’s ability to predict the correct token given the prior context:
\begin{eqnarray}
\begin{aligned}
&\mathcal{L}_{\text{language}} = -\frac{1}{N} \sum_{i=1}^N \log P(y_{\text{label}} | y_{\text{pred}})
\end{aligned}
\end{eqnarray}
where $P(y_{\text{label}} | y_{\text{pred})}$ represents the probability assigned to the correct token $y_{\text{label}}$ by the model, $y_{\text{pred}}$ is the predicted token, and $N$ is the total number of tokens. By minimizing this loss, the model learns to generate coherent textual descriptions of video content, laying the foundation for subsequent stages.

\paragraph{(2) Refining quality scoring with L1 loss.}
Once the model can produce coherent descriptions of video content, the focus shifts to fine-tuning the quality regression module to output stable and precise numerical quality scores. The quality regression module takes the aligned visual tokens as input and predicts a quality score that reflects the overall video quality.
Using the AIGVQA-DB, which contains human-annotated MOS for each video, the model is trained to align its predictions with human ratings. The training objective minimizes the difference between the predicted quality score $Q_{predict}$ and the ground-truth MOS  $Q_{label}$ using the L1 loss function:
\begin{eqnarray}
\begin{aligned}
\label{loss_function}
\mathcal{L}_{\text{MOS}} = \frac{1}{N} \sum_{i=1}^N \left| Q_{\text{predict}}(i) - Q_{\text{label}}(i) \right|
\end{aligned}
\end{eqnarray}
where $Q_{predict}(i)$ is the score predicted by the regressor $i$ and $Q_{label}(i)$ is the corresponding ground-truth MOS derived from subjective experiments, and $N$  is the number of videos in the batch. This loss function ensures that the predicted scores remain consistent with human evaluations, enabling the model to accurately assess the quality of AI-generated videos in numerical form.

\paragraph{(3) Enhancing pairwise comparisons with cross-entropy Loss.}
The third stage incorporates the AIGVQA-DB subset into the training pipeline. This dataset contains human annotations for pairwise video comparisons, where two videos are evaluated, and the superior one is selected based on quality. Pairwise training helps the model learn relative quality distinctions, enabling it to compare videos effectively.
The objective in this stage is to maximize the probability that the model predicts a higher score for the better video in a pair.
The pairwise comparison loss is calculated by comparing the predicted scores for a video pair, which are processed through an LPIPS network to judge which video is better. This predicted logit is then compared with the ground-truth logit labels (0 or 1) using the cross-entropy loss. The label 0 indicates that video2 is better, and 1 indicates that video1 is better. The order of the pair (which video is considered as video1 or video2) is random, but the logit label always corresponds correctly to the better video. 
\begin{eqnarray}
\begin{aligned}
\label{pair_loss}
\mathcal{L}_{\text{Pairs}} = -\frac{1}{N} \sum_{i=1}^N \big[ y_{\text{label}}(i) \log y_{\text{pred}}(i) \\ 
 + (1 - y_{\text{label}}(i)) \log (1 - y_{\text{pred}}(i)) \big]
\end{aligned}
\end{eqnarray}
where $y_{\text{pred}}$ is the logit predicted by the network for the video pair, $y_{\text{label}}$ is the ground-truth label for the video pair (0 for video2 better, 1 for video1 better), and $N$  is the number of videos in the batch.
This function encourages the model to predict the better video in a pair, reinforcing its ability to make accurate comparisons. By incorporating the pairwise data into training, the model not only learns to provide accurate quality scores but also becomes proficient in comparing videos and selecting the superior one. This enhances its utility in real-world applications, where users often need to compare the quality of multiple videos directly.

%% file: supple/4_Experiment.tex
\section{Implemention Details}

\subsection{Detailed Information of Evaluation Criteria}
\noindent
We adopt the widely used metrics in VQA literature \cite{chikkerur2011objective, parmar2019and}: Spearman rank-order correlation coefficient (SRCC), Pearson linear correlation coefficient (PLCC), and Kendall’s Rank Correlation Coefficient (KRCC) as our evaluation criteria. SRCC quantifies the extent to which the ranks of two variables are related, which ranges from -1 to 1. Given $N$ action videos, SRCC is computed as:
\begin{equation}
SRCC = 1 - \frac{{6\sum\nolimits_{n = 1}^N {{{({v_n} - {p_n})}^2}} }}{{N({N^2} - 1)}},
\end{equation}
where $v_n$ and $p_n$ denote the rank of the ground truth $y_n$ and the rank of predicted score ${\hat y_n}$ respectively. The higher the SRCC, the higher the monotonic correlation between ground truth and predicted score.
Similarly, PLCC measures the linear correlation between predicted scores and ground truth scores, which can be formulated as:
\begin{equation}
PLCC = \frac{{\sum\nolimits_{n = 1}^N {({y_n} - \bar y)({{\hat y}_n} - \bar {\hat y})} }}{{\sqrt {\sum\nolimits_{n = 1}^N {{{({y_n} - \bar y)}^2}} } \sqrt {\sum\nolimits_{n = 1}^N {{{({{\hat y}_n} - \bar {\hat y})}^2}} } }},
\end{equation}
where $\bar y$ and $\bar {\hat y}$ are the mean of ground truth and predicted score respectively.
\noindent
We also adopt the Kendall Rank Correlation Coefficient (KRCC) as an evaluation metric, which measures the ordinal association between two variables. For a pair of ranks $(v_i, p_i)$ and $(v_j, p_j)$, the pair is concordant if:
\begin{equation}
(v_i - v_j)(p_i - p_j) > 0,
\end{equation}
and discordant if $< $ 0.
 Given $N$ AIGVs, KRCC is computed as:
\begin{equation}
KRCC =  \frac{{C - D}}{{\frac{1}{2}N(N-1)}},
\end{equation}
where $C$ and $D$ denote the number of concordant and discordant pairs, respectively.

\subsection{Detailed Information of Evaluation Algorithms}
\noindent
\textbf{V-Dynamic}~\cite{huang2024vbench} and \textbf{V-Smoothness}~\cite{huang2024vbench}  are proposed in VBench~\cite{huang2024vbench}. We directly used the respective implementation code in VBench~\cite{huang2024vbench} without specific changes. 

\noindent
{\bf CLIPScore} \cite{CLIPScore} is an image captioning metric, which is widely used to evaluate T2I/T2V models. It passes both the image and the candidate caption through their respective feature extractors, then computing the cosine similarity  between the text and image embeddings.
{\bf BLIPScore} \cite{li2022blip} replace CLIP with BLIP \cite{li2022blip} to compute the cosine similarity. {\bf ImageReward} uses BLIP \cite{li2022blip} as the backbone, and uses an MLP to generate a scalar for preference comparison.
{\bf AestheticScore} is given by an aesthetic predictor introduced by LAION \cite{schuhmann2022laion}. 
 

\noindent
{\bf VSFA}~\cite{VSFA} is an objective no-reference video quality assessment method by integrating two eminent effects of the human visual system, namely, content-dependency and temporal-memory effects into a deep neural network. We directly used the official code without specific changes. 

\noindent
{\bf BVQA}~\cite{li2022blindly} leverages the transferred knowledge from IQA databases with authentic distortions and large-scale action recognition with rich motion patterns for better video representation. We used the officially pre-trained model under mixed-database settings and finetuned it on our AIGVQA-DB for evaluation.

\noindent
{\bf SimpleVQA}~\cite{sun2022a} adopts an end-to-end spatial feature extraction network to directly learn the quality-aware spatial feature representation from raw pixels of the video frames and extract the motion features to measure the temporal-related distortions. A pre-trained SlowFast model is used to extract motion features. We used the officially pre-trained model and finetuned it on our AIGVQA-DB for evaluation.

\noindent
{\bf FAST-VQA}~\cite{wu2022fast} proposes a grid mini-patch sampling strategy, which allows consideration of local quality by sampling patches at their raw resolution and covers global quality with contextual relations via mini-patches sampled in uniform grids. It overcomes the high computational costs when evaluating high-resolution videos. We used the officially released FAST-VQA-B model and finetuned it on our AIGVQA-DB.

\noindent
{\bf DOVER}~\cite{wu2023dover} is a disentangled objective video quality evaluator that learns the quality of videos based on technical and aesthetic perspectives. We used the officially pre-trained model and finetuned it on our AIGVQA-DB.

\noindent
{\bf Q-Align} \cite{wu2023q} is a human-emulating syllabus designed to train large multimodal models for visual scoring tasks. It mimics the process of training human annotators by converting MOS into five text-defined rating levels. We used the officially pre-trained model and finetuned it on our AIGVQA-DB.

\input{table/prompts}

%% file: table/prompts.tex
\begin{table*}
\centering
  \caption{Examples of prompts and their corresponding categorizations in AIGVQA-DB.}
  \label{prompts}
  \vspace{-1mm}
  \renewcommand\arraystretch{1.3}
  \resizebox{\textwidth}{!}{\begin{tabular}{llllll}
    \toprule
      \textbf{Prompts} & \textbf{Spatial major content } &\textbf{Temporal major content } &\textbf{Attribute control}   & \textbf{Complexity} & \textbf{Source} \\
    \midrule 
    “A person is running backwards.”
    & people & actions, kinetic motions & motion direction & simple & FETV \cite{liu2024fetv}\\
    “A plane is flying backwards.”
    & vehicles & kinetic motions & motion direction & simple & FETV \cite{liu2024fetv}\\
    “A blue horse is running in the field.”
    & animals & actions, kinetic motions & color & simple & FETV \cite{liu2024fetv}\\
    “A green shark is swimming under the water.”
    & animals & fluid motions, actions, kinetic motions & color & simple & FETV \cite{liu2024fetv}\\
    “A leave is flying towards the tree from the ground.”
    & plants & kinetic motions & motion direction & medium & FETV \cite{liu2024fetv}\\
     “The flowers first wilt and then bloom again.”
    & plants & fluid motions & event order  & simple & FETV \cite{liu2024fetv}\\
     “The sun sets on the horizon and then immediately rises again.”
    & scenery and natural objects & kinetic motions & event order & medium & FETV \cite{liu2024fetv}\\
     “A person pours a cup of coffee from a bottle and then pours the & people, food and beverage & actions, fluid motions & event order & complex & FETV \cite{liu2024fetv}\\ coffee back to the bottle.”
    \\
    “The three singers are dancing in swim suits.”
    & people & actions & quantity & simple & InternVid \cite{wang2023internvid}\\
    “a bearded man nods and blows kisses.”
    & people & actions & event order & simple & InternVid \cite{wang2023internvid}\\
      “A a dog are flipping and riding a skateboard.”
    & animals & actions & null & simple & InternVid \cite{wang2023internvid}\\
    “A white puppy plays with a slice of lime”
    & animals, food and beverage & kinetic motions & color & medium & InternVid \cite{wang2023internvid}\\
    “Rain is falling on a black umbrella.”
    & plants, scenery and natural object & fluid motions & color & simple &InternVid \cite{wang2023internvid}\\
     “Two cars are racing on a track.”
    & vehicles & kinetic motions & quantity & simple & InternVid \cite{wang2023internvid}\\
     “Two very handsome boys are singing on the stage.”
    & people & actions & quantity & medium & InternVid \cite{wang2023internvid}\\
       “Two girls are standing in the ocean when they become frightened & people & actions & quantity, event order & complex & InternVid \cite{wang2023internvid}\\
    of something in the water.”
    \\
     “An arial view of animals running”
    & animals & actions, kinetic motions & camera view & simple & MSRVTT~\cite{xu2016msr-vtt}\\
     “Overhead view as pingpong players compete on the table”
    & people & actions, kinetic motions & camera view & medium & MSRVTT~\cite{xu2016msr-vtt}\\

     “There is a orange color fish floating in the water”
    & animals & fluid motions & color & medium & MSRVTT~\cite{xu2016msr-vtt}\\
     “Some blue water in a pool is rippling around”
    & scenery and natural objects & fluid motions & color & medium & MSRVTT~\cite{xu2016msr-vtt}\\
    “A red sport car is driving very fast”
    & vehicles & kinetic motions & color, speed & medium & MSRVTT~\cite{xu2016msr-vtt}\\
      “Four friends are driving in the car”
    & scenery and natural objects & fluid motions & color & medium & MSRVTT~\cite{xu2016msr-vtt}\\
    “Smoke is coming out of a mountain”
    & scenery and natural objects & fluid motions & motion direction & simple & MSRVTT~\cite{xu2016msr-vtt}\\
        “Satellite view of moon we can also see sunlight but surface is & scenery and natural objects & light change & camera view & complex & MSRVTT~\cite{xu2016msr-vtt}\\not smooth”\\
    “Smoke billows from the factory chimney.”
    & vehicles, buildings and infrastructure & fluid motions & color & simple & 
    Handwritten\\
    “Leaves flutter from the trees in the gusty wind.”
    & plants & kinetic motions & motion direction & medium & Handwritten\\
    “The crimson hues painted the horizon during the beach sunset.”
    & scenery and natural objects & light change & color & medium & 
    Handwritten\\
    “The static view of a solar eclipse revealed nature's cosmic spectacle.”
    & scenery and natural objects & light change & camera view & medium & Handwritten\\
  
    “A hiker reaches the summit and then admires the breathtaking view.”
    & people & actions & event order & medium & Handwritten\\
    “Vinegar drizzling onto a salad, filmed in intricate detail.”
    & food and beverage & fluid motions & camera view & medium & 
    Handwritten\\
    “An egg cracking open and being whisked vigorously in slow motion.”
    & food and beverage & kinetic motions & speed & medium & Handwritten\\
      “The coastline transformed into a canvas of fiery colors during the   & scenery and natural objects  & light change & color & complex & 
    Handwritten\\beach sunset.”
   \\
     “Two men playing musical instruments in a city square.”
    & people, artifacts & kinetic motions & quantity & medium & TGIF~\cite{li2016tgif}\\
     “The bridge of a river being viewed from a cable.”
    & scenery and natural object & kinetic motions & camera view & medium & TGIF~\cite{li2016tgif}\\
    “A view from inside of a bus showing snow”
    & vehicles, scenery and natural object & kinetic motions & camera view & medium & TGIF~\cite{li2016tgif}\\
     “Some large metal barrels on a train track.”
    & vehicles, artifacts & actions & quantity & simple & TGIF~\cite{li2016tgif}\\
     “A person reaching into a dish of beef and vegetables.”
    & people, food and beverage & actions & quantity & medium & TGIF~\cite{li2016tgif}\\
    “A man wearing a green jacket is fixing a solar panel.”
    & people & actions & color & medium & TGIF~\cite{li2016tgif}\\
     “A green toy chamelon eating a cookie.”
    & animals, food and beverage & kinetic motions & color & simple & TGIF~\cite{li2016tgif}\\
    
    “Two people sit on a table with headphones on”
    & people & actions & quantity & medium & TGIF~\cite{li2016tgif}\\
    “A screenshot of the dashboard of a korean language software”
    & illustrations & actions & camera view & medium & TGIF~\cite{li2016tgif}\\

     “A woman's tight pants with two photos, one showing her wearing & people, artifacts & actions & quantity & complex& TGIF~\cite{li2016tgif}\\  the pants and the other showing her with a shirt.”
    \\
     “Background - sunset landscape beach.”
    & scenery and natural object & light change & null & simple & WebVid \cite{Bain21}\\
     “The musician plays the guitar. close up.”
    & people, artifacts & actions & camera view & simple & WebVid \cite{Bain21}\\
     “Background - sunset landscape beach.”
    & scenery and natural object & light change & null & simple & WebVid \cite{Bain21}\\
     “Background - sunset landscape beach.”
    & scenery and natural object & light change & null & simple & WebVid \cite{Bain21}\\
     “Background - sunset landscape beach.”
    & scenery and natural object & light change & null & simple & WebVid \cite{Bain21}\\
     “Attractive young woman silhouette dancing outdoors on a sunset with  & people, scenery and natural object & actions, light change & speed & complex & WebVid \cite{Bain21}\\sun shining bright behind her on a horizon. slow motion.”
   \\
     “Night landscape timelapse with colorful milky way. starry sky with & plants, scenery and natural object & light change & speed & complex & WebVid \cite{Bain21} \\tropical palms on the island. milky way timelapse over palms.”
    \\
     “A blue wave of fire grows into a large flame and bright sparks on a  & scenery and natural object & fluid motions, light change & camera view, color, speed & complex & WebVid \cite{Bain21}\\shiny surface. closeup. slow motion, high speed camera.”
   \\
     “Silhouette of happy mom dad and baby at sunset in a field with wheat. \\farmer and family on the field. a child with parents plays in the wheat.   & people, plants & actions, light change & null & complex & WebVid \cite{Bain21} \\the concept of family relationships.”
 \\
   “Traditional chinese ink preparation. low angle dolly shot close up \\focus from brushes on ceramic stand to person hands in background  & artifacts, people & actions & camera view & complex & WebVid \cite{Bain21} \\preparing ink for calligraphy.”
   \\
     “Beautiful growing network with economic indicators growing abstract\\  seamless. looped 3d animation of moving numbers and lines. cyberspace& illustrations & light change & camera view & complex & WebVid \cite{Bain21} \\ flashing lights. business concept. 4k ultra hd 3840x2160.”
    \\

    \bottomrule
  \end{tabular}}
    \vspace{-3mm}
\end{table*}

%% file: supple/5_figures.tex
\input{figure/sup_smc}
\input{figure/sup_tmc}
\input{figure/sup_attri}
\input{figure/sup_bz1}
\input{figure/sup_bz2}
\input{figure/sup_bz3}
\input{figure/sup_bz4}

%% file: figure/sup_smc.tex
\begin{figure*}[!t]
	\centering
	\includegraphics[width=\linewidth]{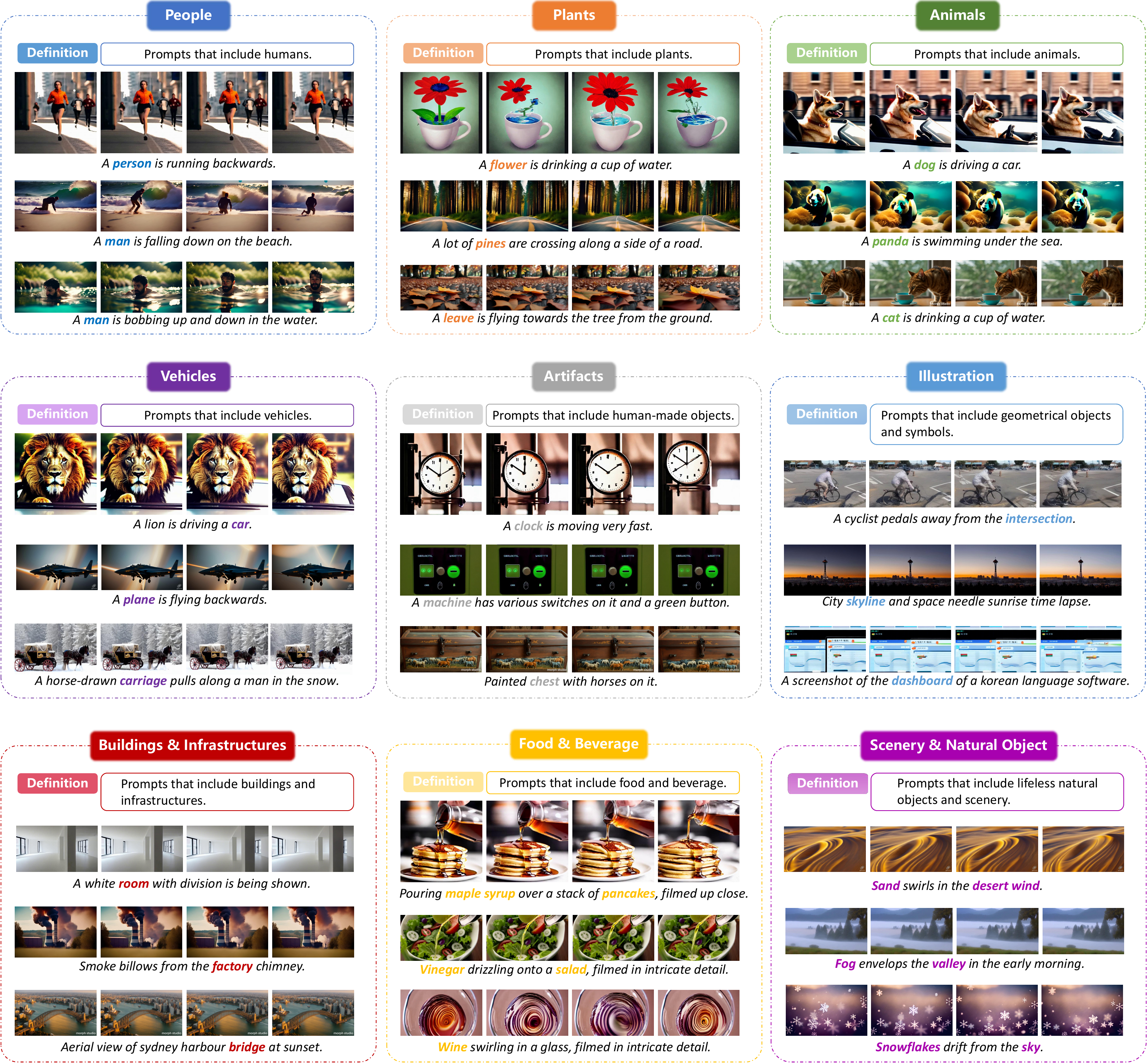}
 \vspace{-5mm}
	\caption{Descriptions and examples of the spatial major contents.}
 \vspace{-3mm}
	\label{sup_mos1}
\end{figure*}

%% file: figure/sup_tmc.tex
\begin{figure*}[!t]
	\centering
	\includegraphics[width=\linewidth]{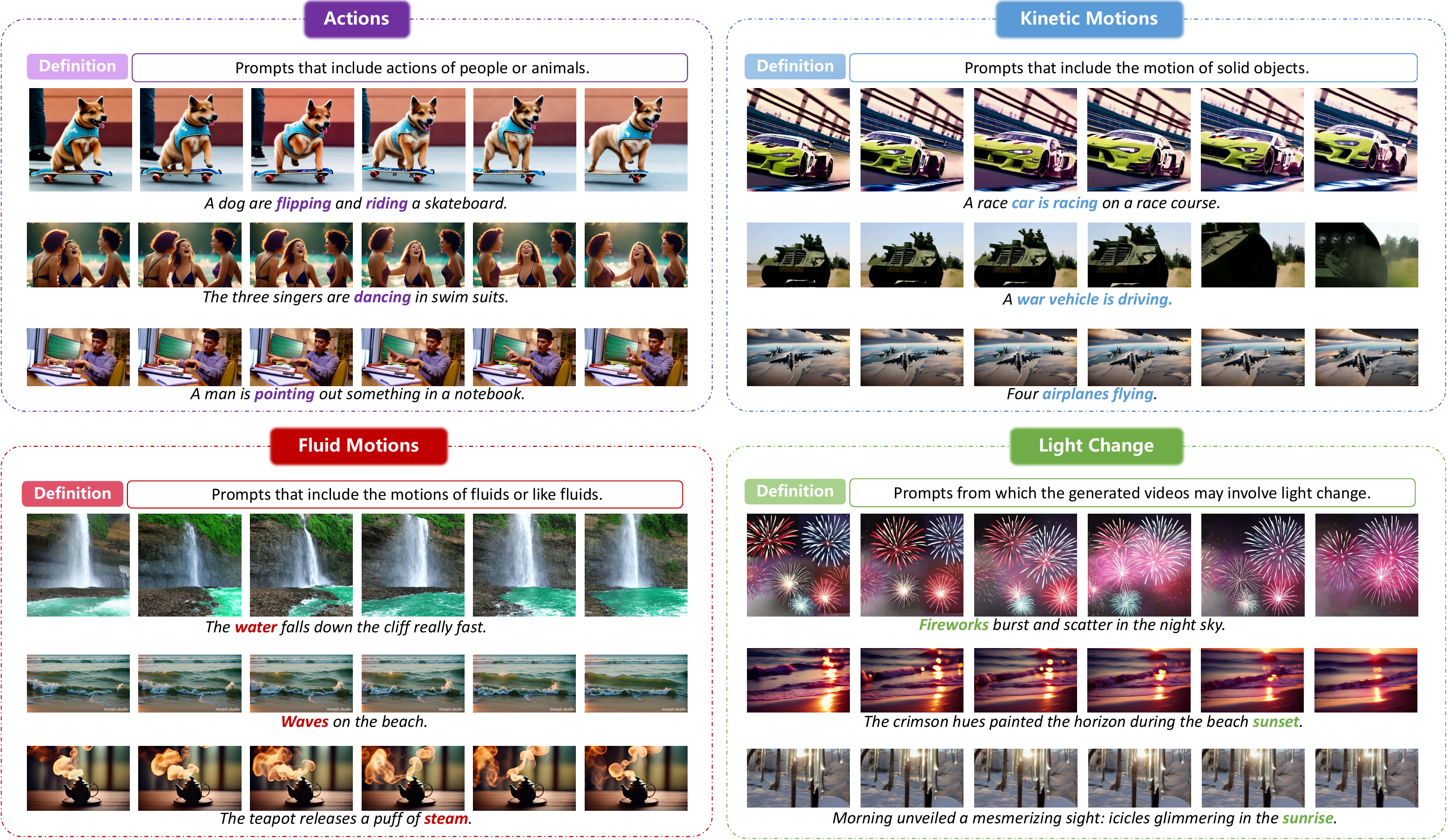}
 \vspace{-5mm}
	\caption{Descriptions and examples of the temporal major contents.}
 \vspace{-3mm}
	\label{sup_mos1}
\end{figure*}

%% file: figure/sup_attri.tex
\begin{figure*}[!t]
	\centering
	\includegraphics[width=\linewidth]{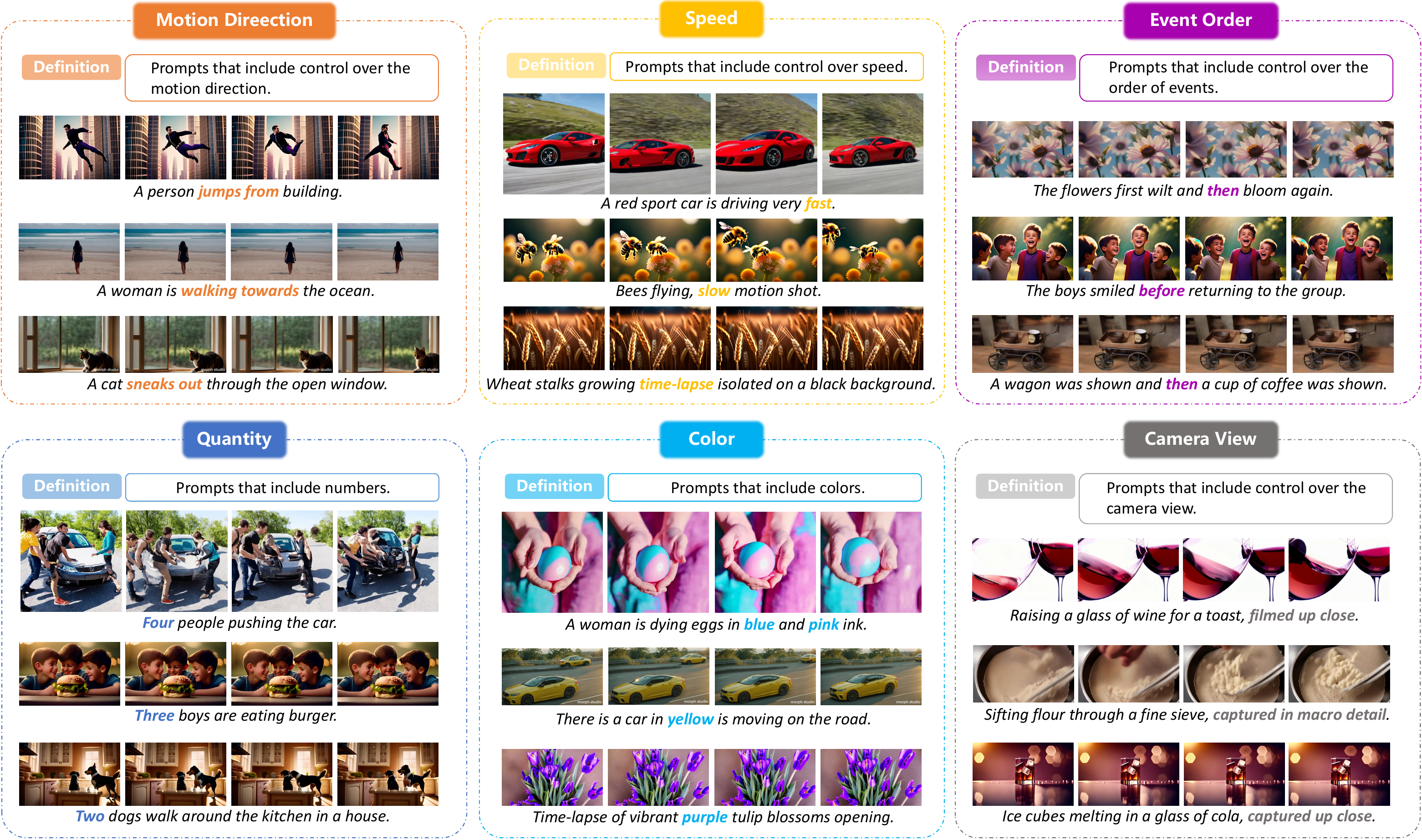}
 \vspace{-5mm}
	\caption{Descriptions and examples of the attribute control.}
 \vspace{-3mm}
	\label{sup_mos1}
\end{figure*}

%% file: figure/sup_bz1.tex
\begin{figure*}[!t]
\vspace{-3mm}
	\centering
	\includegraphics[width=\linewidth]{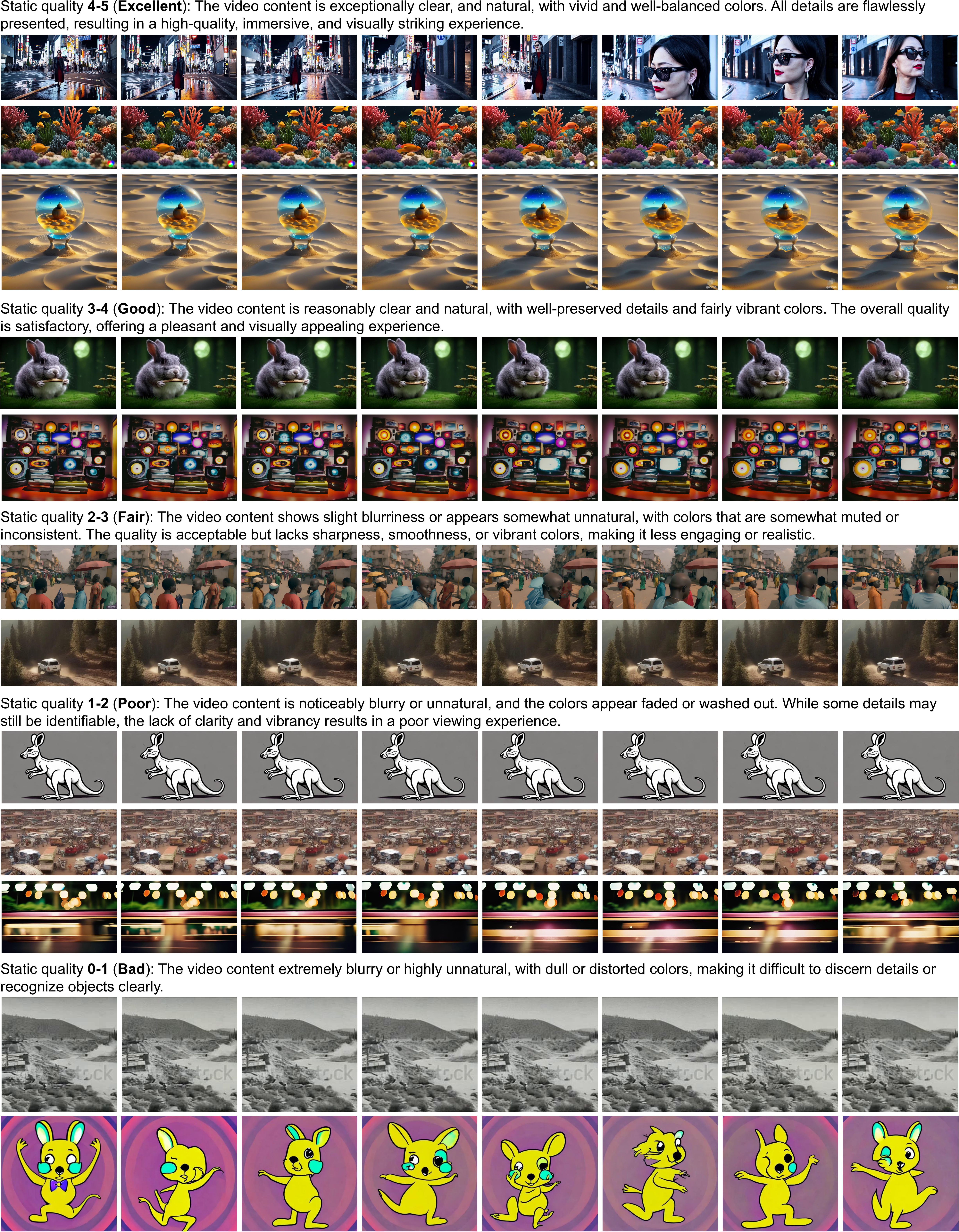}
 \vspace{-5mm}
	\caption{Instructions and examples for manual evaluation of \textbf{static quality}.}
 \vspace{-3mm}
	\label{sup_bz1}
\end{figure*}

%% file: figure/sup_bz2.tex
\begin{figure*}[!t]
\vspace{-3mm}
	\centering
	\includegraphics[width=\linewidth]{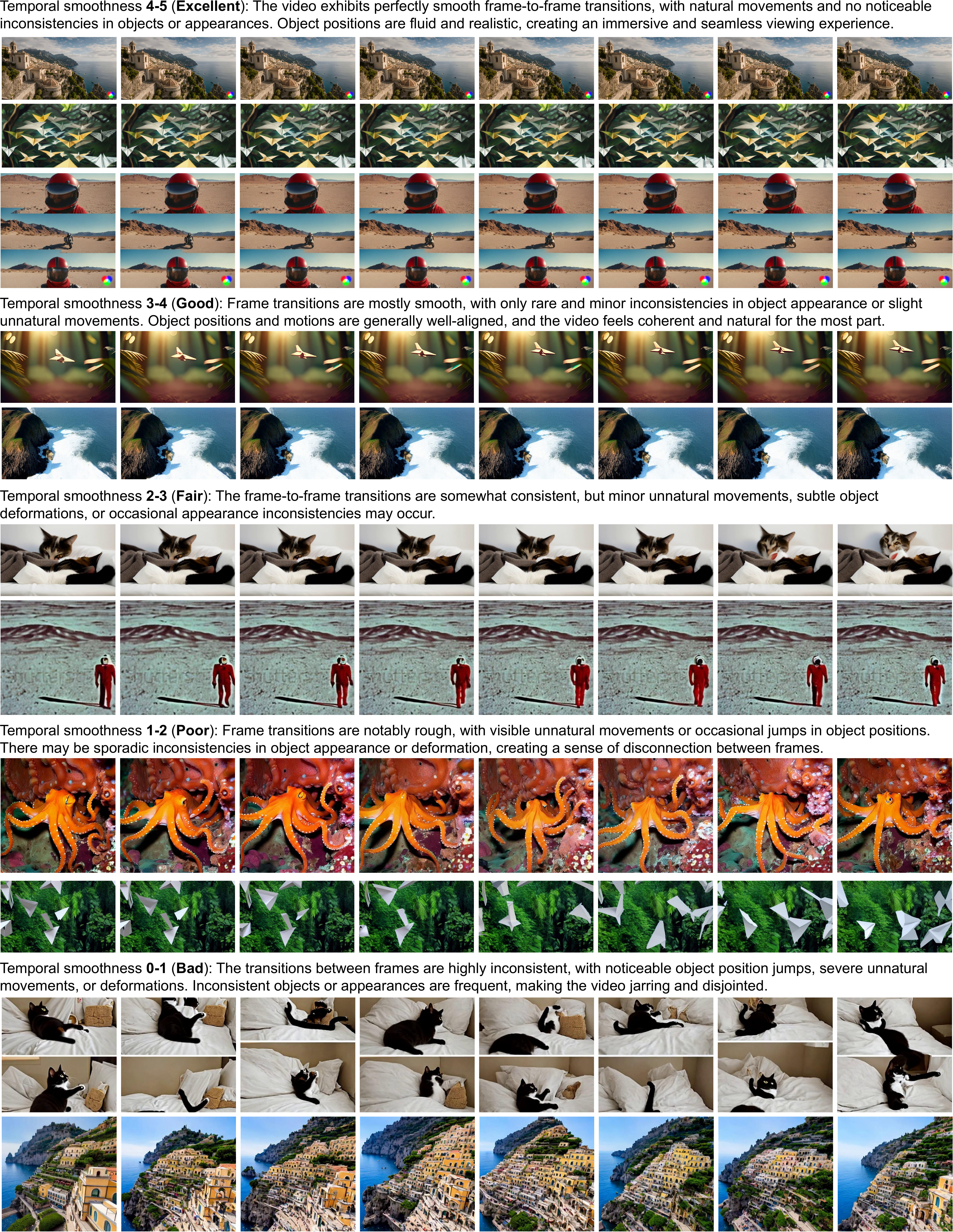}
 \vspace{-5mm}
	\caption{Instructions and examples for manual evaluation of \textbf{temporal smoothness}.}
 \vspace{-3mm}
	\label{sup_bz2}
\end{figure*}

%% file: figure/sup_bz3.tex
\begin{figure*}[!t]
\vspace{-3mm}
	\centering
	\includegraphics[width=\linewidth]{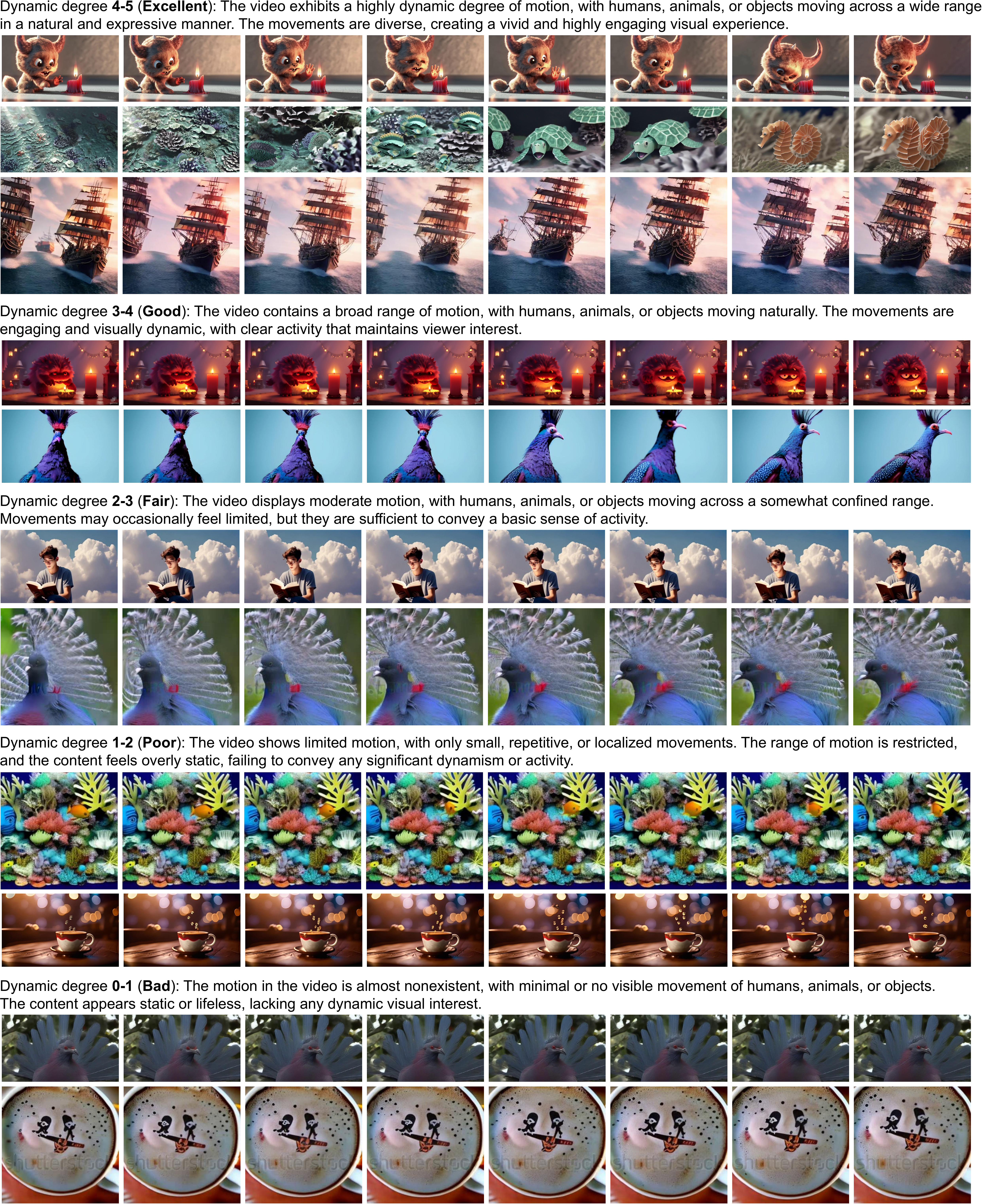}
 \vspace{-5mm}
	\caption{Instructions and examples for manual evaluation of \textbf{dynamic degree}.}
 \vspace{-3mm}
	\label{sup_bz3}
\end{figure*}

%% file: figure/sup_bz4.tex
\begin{figure*}[!t]
\vspace{-3mm}
	\centering
	\includegraphics[width=\linewidth]{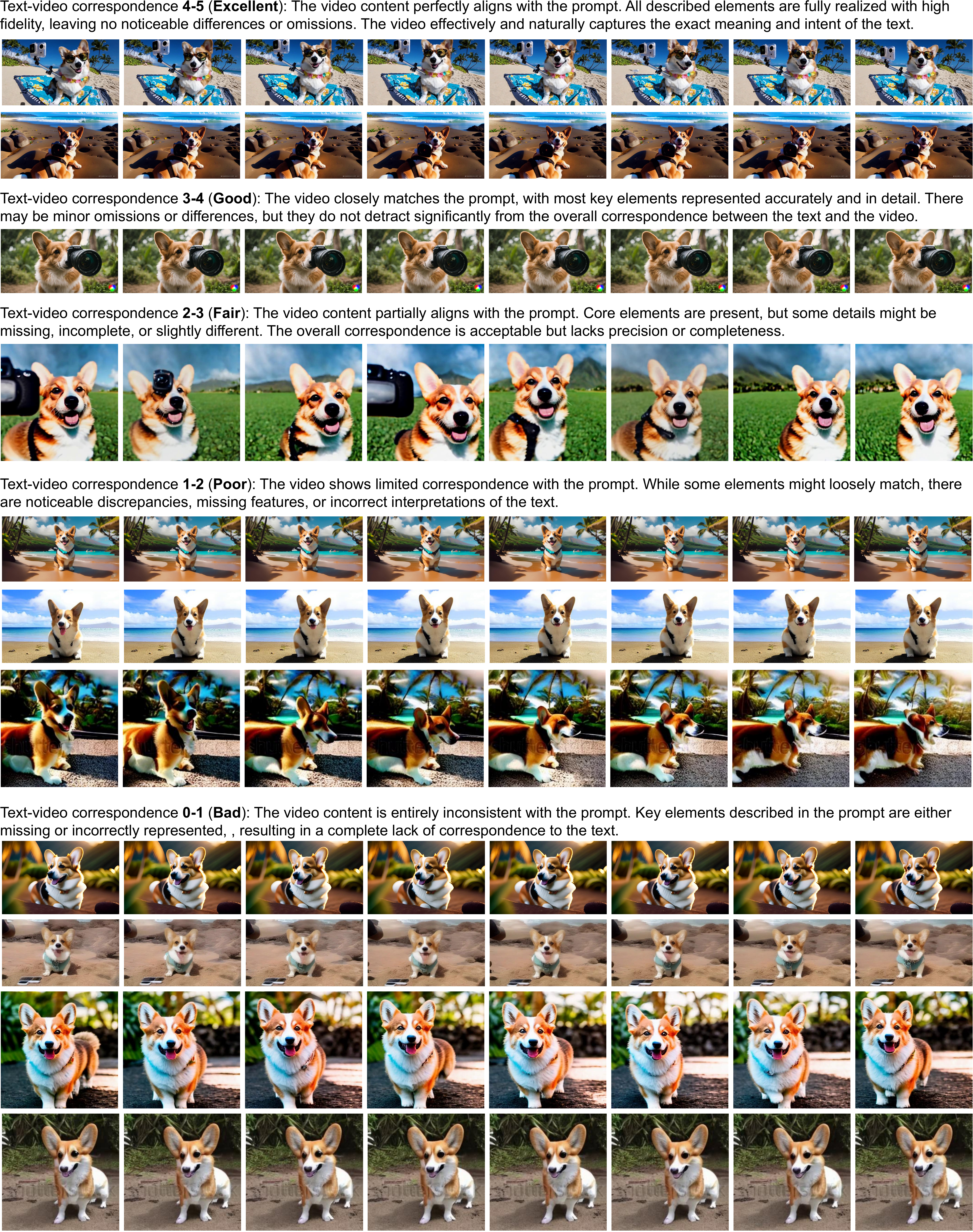}
 \vspace{-5mm}
	\caption{Instructions and examples for manual evaluation of \textbf{text-video correspondence}. The example videos are of the same prompt “A corgi vlogging itself in tropical Maui.”}
 \vspace{-3mm}
	\label{sup_bz4}
\end{figure*}